# Best-First Heuristic Search for Multicore Machines

**Ethan Burns**                                                    EABURNS AT CS.UNH.EDU
**Sofia Lemons**                                               SOFIA.LEMONS AT CS UNH.EDU
**Wheeler Ruml**                                                     RUML AT CS.UNH.EDU
*Department of Computer Science*
*University of New Hampshire*
*Durham, NH 03824 USA*

**Rong Zhou**                                                       RZHOU AT PARC.COM
*Embedded Reasoning Area*
*Palo Alto Research Center*
*Palo Alto, CA 94304 USA*

## Abstract

To harness modern multicore processors, it is imperative to develop parallel versions of fundamental algorithms. In this paper, we compare different approaches to parallel best-first search in a shared-memory setting. We present a new method, PBNF, that uses abstraction to partition the state space and to detect duplicate states without requiring frequent locking. PBNF allows speculative expansions when necessary to keep threads busy. We identify and fix potential livelock conditions in our approach, proving its correctness using temporal logic. Our approach is general, allowing it to extend easily to suboptimal and anytime heuristic search. In an empirical comparison on STRIPS planning, grid pathfinding, and sliding tile puzzle problems using 8-core machines, we show that A*, weighted A* and Anytime weighted A* implemented using PBNF yield faster search than improved versions of previous parallel search proposals.

## 1. Introduction

It is widely anticipated that future microprocessors will not have faster clock rates, but instead more computing cores per chip. Tasks for which there do not exist effective parallel algorithms will suffer a slowdown relative to total system performance. In artificial intelligence, heuristic search is a fundamental and widely-used problem solving framework. In this paper, we compare different approaches for parallelizing best-first search, a popular method underlying algorithms such as Dijkstra's algorithm and A* (Hart, Nilsson, & Raphael, 1968).

In best-first search, two sets of nodes are maintained: *open* and *closed*. Open contains the search frontier: nodes that have been generated but not yet expanded. In A*, open nodes are sorted by their $f$ value, the estimated lowest cost for a solution path going through that node. Open is typically implemented using a priority queue. Closed contains all previously generated nodes, allowing the search to detect states that can be reached via multiple paths in the search space and avoid expanding them multiple times. The closed list is typically implemented as a hash table. The central challenge in parallelizing best-first search is avoiding contention between threads when accessing the open and closed lists. We look at a variety of methods for parallelizing best-first search, focusing on algorithms which are based on two techniques: *parallel structured duplicate detection* and *parallel retracting A*.

Parallel structured duplicate detection (PSDD) was originally developed by Zhou and Hansen (2007) for parallel breadth-first search, in order to reduce contention on shared data structures by allowing threads to enjoy periods of synchronization-free search. PSDD requires the user to supply an abstraction function that maps multiple states, called an *n*block, to a single abstract state. We present a new algorithm based on PSDD called Parallel Best-*N* Block-First (PBNF[1]). Unlike PSDD, PBNF extends easily to domains with non-uniform and non-integer move costs and inadmissible heuristics. Using PBNF in an infinite search space can give rise to livelock, where threads continue to search but a goal is never expanded. We will discuss how this condition can be avoided in PBNF using a method we call *hot nblocks*, as well as our use of bounded model checking to test its effectiveness. In addition, we provide a proof of correctness for the PBNF framework, showing its liveness and completeness in the general case.

Parallel retracting A* (PRA*) was created by Evett, Hendler, Mahanti, and Nau (1995). PRA* distributes the search space among threads by using a hash of a node's state. In PRA*, duplicate detection is performed locally; communication with peers is only required to transfer generated search-nodes to their home processor. PRA* is sensitive to the choice of hashing function used to distribute the search space. We show a new hashing function, based on the same state space abstraction used in PSDD, that can give PRA* significantly better performance in some domains. Additionally, we show that the communication cost incurred in a naive implementation of PRA* can be prohibitively expensive. Kishimoto, Fukunaga, and Botea (2009) present a method that helps to alleviate the cost of communication in PRA* by using asynchronous message passing primitives.

We evaluate PRA* (and its variants), PBNF and other algorithms empirically using dual quad-core Intel machines. We study their behavior on three popular search domains: STRIPS planning, grid pathfinding, and the venerable sliding tile puzzle. Our empirical results show that the simplest parallel search algorithms are easily outperformed by a serial A* search even when they are run with eight threads. The results also indicate that adding abstraction to the PRA* algorithm can give a larger increase in performance than simply using asynchronous communication, although using both of these modifications together may outperform either one used on its own. Overall, the PBNF algorithm often gives the best performance.

In addition to finding optimal solutions, we show how to adapt several of the algorithms to bounded suboptimal search, quickly finding $w$-admissible solutions (with cost within a factor of $w$ of optimal). We provide new pruning criteria for parallel suboptimal search and prove that algorithms using them retain $w$-admissibility. Our results show that, for sufficiently difficult problems, parallel search may significantly outperform serial weighted A* search. We also found that the advantage of parallel suboptimal search increases with problem difficulty.

Finally, we demonstrate how some parallel searches, such as PBNF and PRA*, lead naturally to effective anytime algorithms. We also evaluate other obvious parallel anytime search strategies such as running multiple weighted A* searches in parallel with different weights. We show that the parallel anytime searches are able to find better solutions faster than their serial counterparts and they are also able to converge more quickly on optimal solutions.

---

1. Peanut Butter 'N' (marshmallow) Fluff, also known as a fluffernutter, is a well-known children's sandwich in the USA.

## 2. Previous Approaches

There has been much previous work in parallel search. We will briefly summarize selected proposals before turning to the foundation of our work, the PRA* and PSDD algorithms.

### 2.1 Depth- and Breadth-first Approaches

Early work on parallel heuristic search investigated approaches based on depth-first search. Two examples are distributed tree search (Ferguson & Korf, 1988), and parallel window search (Powley & Korf, 1991).

Distributed tree search begins with a single thread, which is given the initial state to expand. Each time a node is generated an unused thread is assigned to the node. The threads are allocated down the tree in a depth-first manner until there are no more free threads to assign. When this occurs, each thread will continue searching its own children with a depth-first search. When the solution for a subtree is found it is passed up the tree to the parent thread and the child thread becomes free to be re-allocated elsewhere in the tree. Parent threads go to sleep while their children search, only waking once the children terminate, passing solutions upward to their parents recursively. Because it does not keep a closed list, depth-first search cannot detect duplicate states and does not give good search performance on domains with many duplicate states, such as grid pathfinding and some planning domains.

Parallel window search parallelizes the iterative deepening A* (IDA*, see Korf, 1985) algorithm. In parallel window search, each thread is assigned a cost-bound and will perform a cost-bounded depth-first search of the search space. The problem with this approach is that IDA* will spend at least half of its search time on the final iteration and since every iteration is still performed in only a single thread, the search will be limited by the speed of a single thread. In addition, non-uniform costs can foil iterative deepening, because there may not be a good way to choose new upper-bounds that give the search a geometric growth.

Holzmann and Bosnacki (2007) have been able to successfully parallelize depth-first search for model checking. The authors are able to demonstrate that their technique that distributes nodes based on search depth was able to achieve near linear speedup in the domain of model checking. Other research has used graphics processing units (GPUs) to parallelize breadth-first search for use in two-player games (Edelkamp & Sulewski, 2010). In the following sections we describe algorithms with the intent of parallelizing best-first search.

### 2.2 Simple Parallel Best-first Search

The simplest approach to parallel best-first search is to have open and closed lists that are shared among all threads (Kumar, Ramesh, & Rao, 1988). To maintain consistency of these data structures, mutual exclusion locks (mutexes) need to be used to ensure that a single thread accesses the data structure at a time. We call this search "parallel A*". Since each node that is expanded is taken from the open list and each node that is generated is looked up in the closed list by every thread, this approach requires a lot of synchronization overhead to ensure the consistency of its data structures. As we see in Section 4.3, this naive approach performs worse than serial A*.

There has been much work on designing complex data structures that retain correctness under concurrent access. The idea behind these special *wait-free* data structures is that many threads can use portions of the data structure concurrently without interfering with one another. Most of these approaches use a special *compare-and-swap* primitive to ensure that, while modifying the structure, it does not get modified by another thread. We implemented a simple parallel A* search, which we call lock-free parallel A*, in which all threads access a single shared, concurrent priority queue and concurrent hash table for the open and closed lists, respectively. We implemented the concurrent priority queue data structure of Sundell and Tsigas (2005). For the closed list, we used a concurrent hash table which is implemented as an array of buckets, each of which is a concurrent ordered list as developed by Harris (2001). These lock-free data structures used to implement LPA* require a special lock-free memory manager that uses reference counting and a *compare-and-swap* based stack to implement a free list (Valois, 1995). We will see that, even with these sophisticated structures, a straightforward parallel implementation of A* does not give competitive performance.

One way of avoiding contention altogether is to allow one thread to handle synchronization of the work done by the other threads. $K$-Best-First Search (Felner, Kraus, & Korf, 2003) expands the best $k$ nodes at once, each of which can be handled by a different thread. In our implementation, a master thread takes the $k$ best nodes from open and gives one to each worker. The workers expand their nodes and the master checks the children for duplicates and inserts them into the open list. This allows open and closed to be used without locking, however, in order to adhere to a strict $k$-best-first ordering this approach requires the master thread to wait for all workers to finish their expansions before handing out new nodes. In the domains used in this paper, where node expansion is not particularly slow, we show that this method does not scale well.

One way to reduce contention during search is to access the closed list less frequently. A technique called *delayed duplicate detection* (DDD) (Korf, 2003), originally developed for external-memory search, can be used to temporarily delay access to the a closed list. While several variations have been proposed, the basic principle behind DDD is that generated nodes are added to a single list until a certain condition is met (a depth level is fully expanded, some maximum list size is reached (Stern & Dill, 1998), etc.) Once this condition has been met, the list is sorted to draw duplicate nodes together. All nodes in the list are then checked against the closed list, with only the best version being kept and inserted onto the open list. The initial DDD algorithm used a breadth-first frontier search and therefore only the previous depth-layer was required for duplicate detection. A parallel version was later presented by Niewiadomski, Amaral, and Holte (2006a), which split each depth layer into sections and maintained separate input and output lists for each. These were later merged in order to perform the usual sorting and duplicate detection methods. This large synchronization step, however, will incur costs similar to KBFS. It also depends upon an expensive workload distribution scheme to ensure that all processors have work to do, decreasing the bottleneck effect of nodes being distributed unevenly, but further increasing the algorithm's overhead. A later parallel best-first frontier search based on DDD was presented (Niewiadomski, Amaral, & Holte, 2006b), but incurs even further overhead by requiring synchronization between all threads to maintain a strict best-first ordering.

Jabbar and Edelkamp (2006) present an algorithm called parallel external A* (PEA*) that uses distributed computing nodes and external memory to perform a best-first search. PEA* splits the search space into a set of "buckets" that each contain nodes with the same $g$ and $h$ values. The algorithm performs a best-first search by exploring all the buckets with the lowest $f$ value beginning with the one with the lowest $g$. A master node manages requests to distribute portions of the current bucket to various processing nodes so that expanding a single bucket can be performed in parallel. To avoid contention, PEA* relies on the operating system to synchronize access to files that are shared among all of the nodes. Jabbar and Edelkamp used the PEA* algorithm to parallelize a model-checker and achieved almost linear speedup. While partitioning on $g$ and $h$ works on some domains it is not general if few nodes have the same $g$ and $h$ values. This tends to be the case in domains with real-valued edge costs. We now turn our attention to two algorithms that will reappear throughout the rest of this paper: PRA* and PSDD.

## 2.3 Parallel Retracting A*

PRA* (Evett et al., 1995) attempts to avoid contention by assigning separate open and closed lists to each thread. A hash of the state representation is used to assign nodes to the appropriate thread when they are generated. (Full PRA* also includes a retraction scheme that reduces memory use in exchange for increased computation time; we do not consider that feature in this paper.) The choice of hash function influences the performance of the algorithm, since it determines the way that work is distributed. Note that with standard PRA*, any thread may communicate with any of its peers, so each thread needs a synchronized message queue to which peers can add nodes. In a multicore setting, this is implemented by requiring a thread to take a lock on the message queue. Typically, this requires a thread that is sending (or receiving) a message to wait until the operation is complete before it can continue searching. While this is less of a bottleneck than having a single global, shared open list, we will see below that it can still be expensive. It is also interesting to note that PRA* and the variants mentioned below practice a type of delayed duplicate detection, because they store duplicates temporarily before checking them against a thread-local closed list and possibly inserting them into the open list.

### 2.3.1 IMPROVEMENTS

Kishimoto et al. (2009) note that the original PRA* implementation can be improved by removing the synchronization requirement on the message queues between nodes. Instead, they use the asynchronous send and receive functionality from the MPI message passing library (Snir & Otto, 1998) to implement an asynchronous version of PRA* that they call Hash Distributed A* (HDA*). HDA* distributes nodes using a hash function in the same way as PRA*, except the sending and receiving of nodes happens asynchronously. This means that threads are free to continue searching while nodes which are being communicated between peers are in transit.

In contact with the authors of HDA*, we have created an implementation of HDA* for multicore machines that does not have the extra overhead of message passing for asynchronous communication between threads in a shared memory setting. Also, our implementation of HDA* allows us to make a fair comparison between algorithms by sharing common data structures such as priority queues and hash tables.

In our implementation, each HDA* thread is given a single queue for incoming nodes and one outgoing queue for each peer thread. These queues are implemented as dynamically sized arrays of pointers to search nodes. When generating nodes, a thread performs a non-blocking call to acquire the lock[2] for the appropriate peer's incoming queue, acquiring the lock if it is available and immediately returning failure if it is busy, rather than waiting. If the lock is acquired then a simple pointer copy transfers the search node to the neighboring thread. If the non-blocking call fails the nodes are placed in the outgoing queue for the peer. This operation does not require a lock because the outgoing queue is local to the current thread. After a certain number of expansions, the thread attempts to flush the outgoing queues, but it is never forced to wait on a lock to send nodes. It also attempts to consume its incoming queue and only waits on the lock if its open list is empty, because in that case it has no other work to do. Using this simple and efficient implementation, we confirmed the results of Kishimoto et al. (2009) that show that the asynchronous version of PRA* (called HDA*) outperforms the standard synchronous version. Full results are presented in Section 4.

---

2. One such non-blocking call is the `pthread_mutex_trylock` function of the POSIX standard.

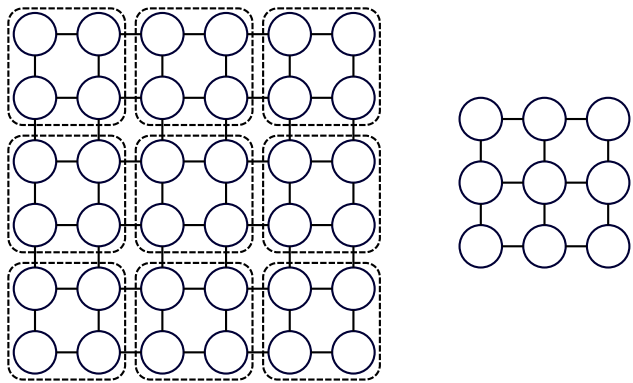

Figure 1: A simple abstraction. Self-loops have been eliminated.

PRA* and HDA* use a simple representation-based node hashing scheme that is the same one, for example, used to look up nodes in closed lists. We present two new variants, APRA* and AHDA*, that make use of state space abstraction to distribute search nodes among the processors. Instead of assigning nodes to each thread, each thread is assigned a set of blocks of the search space where each block corresponds to a state in the abstract space. The intuition behind this approach is that the children of a single node will be assigned to a small subset of all of the remote threads and, in fact, can often be assigned back to the expanding thread itself. This reduces the number of edges in the communication graph among threads during search, reducing the chances for thread contention. Abstract states are distributed evenly among all threads by using a modulus operator in the hope that open nodes will always be available to each thread.

### 2.4 Parallel Structured Duplicate Detection

PSDD is the major previously-proposed alternative to PRA*. The intention of PSDD is to avoid the need to lock on every node generation and to avoid explicitly passing individual nodes between threads. It builds on the idea of structured duplicate detection (SDD), which was originally developed for external memory search (Zhou & Hansen, 2004). SDD uses an *abstraction function*, a many-to-one mapping from states in the original search space to states in an abstract space. The abstract node to which a state is mapped is called its *image*. An $n$block is the set of nodes in the state space that have the same image in the abstract space. The abstraction function creates an *abstract graph* of nodes that are images of the nodes in the state space. If two states are successors in the state space, then their images are successors in the abstract graph. Figure 1 shows a state space graph (left) consisting of 36 nodes and an abstract graph (right) which consists of nine nodes. Each node in the abstract graph represents a grouping of four nodes, called an $n$block, in the original state space, shown by the dotted lines in the state space graph on the left.

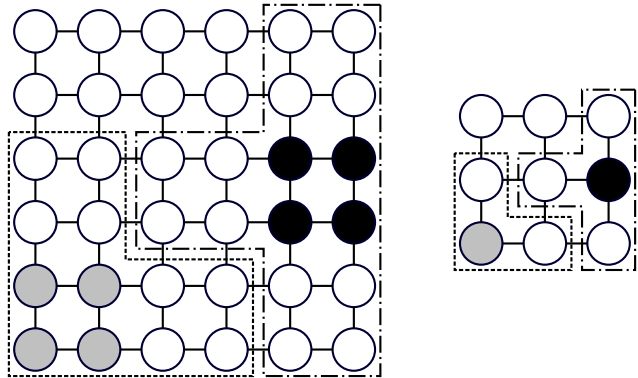

Figure 2: Two disjoint duplicate detection scopes.

Each $n$block has an open and closed list. To avoid contention, a thread will acquire exclusive access to an $n$block. Additionally, the thread acquires exclusive access to the $n$blocks that correspond to the successors in the abstract graph of the $n$block that it is searching. For each $n$block we call the set of $n$blocks that are its successors in the abstract graph its *duplicate detection scope*. This is because these are the only abstract nodes to which access is required in order to perform perfect duplicate detection when expanding nodes from the given $n$block. If a thread expands a node $n$ in $n$block $b$ the children of $n$ must fall within $b$ or one of the $n$blocks that are successors of $b$ in the abstract graph. Threads can determine whether or not new states generated from expanding $n$ are duplicates by simply checking the closed lists of $n$blocks in the duplicate detection scope. This does not require synchronization because the thread has exclusive access to this set of $n$blocks.

In PSDD, the abstract graph is used to find $n$blocks whose duplicate detection scopes are disjoint. These $n$blocks can be searched in parallel without any locking during node expansions. Figure 2 shows two disjoint duplicate detection scopes delineated by dashed lines with different patterns. An $n$block that is not in use by any thread and whose duplicate detection scope is also not in use is considered to be *free*. A free $n$block is available for a thread to acquire it for searching. Free $n$blocks are found by explicitly tracking, for each $n$block $b$, $\sigma(b)$, the number of $n$blocks among $b$'s successors that are in use by another thread. An $n$block $b$ can only be acquired when $\sigma(b) = 0$.

The advantage of PSDD is that it only requires a single lock, the one controlling manipulation of the abstract graph, and the lock only needs to be acquired by threads when finding a new free $n$block to search. This means that threads do not need to synchronize while expanding nodes, their most common operation.

Zhou and Hansen (2007) used PSDD to parallelize breadth-first heuristic search (Zhou & Hansen, 2006). In this algorithm, each $n$block has two lists of open nodes. One list contains open nodes at the current search depth and the other contains nodes at the next search depth. In each thread, only the nodes at the current search depth in an acquired $n$block are expanded. The children that are generated are put in the open list for the next depth in the $n$block to which they map (which will be in the duplicate detection scope of the $n$block being searched) as long as they are not duplicates. When the current $n$block has no more nodes at the current depth, it is swapped for a free $n$block that does have open nodes at this depth. If no more $n$blocks have open nodes at the current depth, all threads synchronize and then progress together to the next depth. An admissible heuristic is used to prune nodes that fall on or above the current solution upper bound.

### 2.4.1 IMPROVEMENTS

While PSDD can be viewed as a general framework for parallel search, in our terminology, PSDD refers to an instance of SDD in a parallel setting that uses layer-based synchronization and breadth-first search. In this subsection, we present two algorithms that use the PSDD framework and attempt to improve on the PSDD algorithm in specific ways.

As implemented by Zhou and Hansen (2007), the PSDD algorithm uses the heuristic estimate of a node only for pruning; this is only effective if a tight upper bound is already available. To cope with situations where a good bound is not available, we have implemented a novel algorithm using the PSDD framework that uses iterative deepening (IDPSDD) to increase the bound. As we report below, this approach is not effective in domains such as grid pathfinding that do not have a geometrically increasing number of nodes within successive $f$ bounds.

Another drawback of PSDD is that breadth-first search cannot guarantee optimality in domains where operators have differing costs. In anticipation of these problems, Zhou and Hansen (2004) suggest two possible extensions to their work, best-first search and a speculative best-first layering approach that allows for larger layers in the cases where there are few nodes (or $n$blocks) with the same $f$ value. To our knowledge, we are the first to implement and test these algorithms.

Best-first PSDD (BFPSDD) uses $f$ value layers instead of depth layers. This means that all nodes that are expanded in a given layer have the same (lowest) $f$ value. BFPSDD provides a best-first search order, but may incur excessive synchronization overhead if there are few nodes in each $f$ layer. To ameliorate this, we loosen the best-first ordering by enforcing that at least $m$ nodes are expanded before abandoning a non-empty $n$block. (Zhou & Hansen, 2007 credit Edelkamp & Schrödl, 2000 with this idea.) Also, when populating the list of free $n$blocks for each layer, all of the $n$blocks that have nodes with the current layer's $f$ value are used or a minimum of $k$ $n$blocks are added where $k$ is four times the number of threads. (This value for $k$ gave better performance than other values tested.) This allows us to add additional $n$blocks to small layers in order to amortize the cost of synchronization. In addition, we tried an alternative implementation of BFPSDD that used a range of $f$ values for each layer. A parameter $\Delta f$ was used to proscribe the width (in $f$ values) of each layer of search. This implementation did not perform as well and we do not present results for it. With either of these enhancements, threads may expand nodes with $f$ values greater than that of the current layer. Because the first solution found may not be optimal, search continues until all remaining nodes are pruned by the incumbent solution.

Having surveyed the existing approaches to parallel best-first search, we now present a new approach which comprises the main algorithmic contribution of this paper.

## 3. Parallel Best-$N$Block-First (PBNF)

In an ideal scenario, all threads would be busy expanding $n$blocks that contain nodes with the lowest $f$ values. To approximate this, we combine PSDD's duplicate detection scopes with an idea from the Localized A* algorithm of Edelkamp and Schrödl (2000). Localized A*, which was designed to improve the locality of external memory search, maintains sets of nodes that reside on the same memory page. The decision of which set to process next is made with the help of a heap of sets

1. while there is an $n$block with open nodes
2.    lock; $b \leftarrow$ best free $n$block; unlock
3.    while $b$ is no worse than the best free $n$block or we've done fewer than *min* expansions
4.        $m \leftarrow$ best open node in $b$
5.        if $f(m) \geq f(incumbent)$, prune all open nodes in $b$
6.        else if $m$ is a goal
7.            if $f(m) < f(incumbent)$
8.                lock; *incumbent* $\leftarrow m$; unlock
9.        else for each child $c$ of $m$
10.            if $c$ is not on the closed list of its $n$block
11.                insert $c$ in the open list of the appropriate $n$block

Figure 3: A sketch of basic PBNF search, showing locking.

ordered by the minimum $f$ value in each set. By maintaining a heap of free $n$blocks ordered on each $n$blocks best $f$ value, we can approximate our ideal parallel search. We call this algorithm Parallel Best-$N$Block-First (PBNF) search.

In PBNF, threads use the heap of free $n$blocks to acquire the free $n$block with the best open node. A thread will search its acquired $n$block as long as it contains nodes that are better than those of the $n$block at the front of the heap. If the acquired $n$block becomes worse than the best free one, the thread will attempt to release its current $n$block and acquire the better one which contains open nodes with lower $f$ values. There is no layer synchronization, so threads do not need to wait unless no $n$blocks are free. The first solution found may be suboptimal, so search must continue until all open nodes have $f$ values worse than the incumbent solution. Figure 3 shows high-level pseudo-code for the algorithm.

Because PBNF is designed to tolerate a search order that is only approximately best-first, we have freedom to introduce optimizations that reduce overhead. It is possible that an $n$block has only a small number of nodes that are better than the best free $n$block, so we avoid excessive switching by requiring a minimum number of expansions. Due to the minimum expansion requirement it is possible that the nodes expanded by a thread are arbitrarily worse than the frontier node with the minimum $f$. We refer to these expansions as "speculative." This can be viewed as trading off node quality for reduced contention on the abstract graph. Section 4.1 shows the results of an experiment that evaluates this trade off.

Our implementation also attempts to reduce the time a thread is forced to wait on a lock by using non-blocking operations to acquire the lock whenever possible. Rather than sleeping if a lock cannot be acquired, a non-blocking lock operation (such as `pthread_mutex_trylock`) will immediately return failure. This allows a thread to continue expanding its current $n$block if the lock is busy. Both of these optimizations can introduce additional 'speculative' expansions that would not have been performed in a serial best-first search.

### 3.1 Livelock

The greedy free-for-all order in which PBNF threads acquire free $n$blocks can lead to livelock in domains with infinite state spaces. Because threads can always acquire new $n$blocks without waiting for all open nodes in a layer to be expanded, it is possible that the $n$block containing the goal will never become free. This is because we have no assurance that all $n$blocks in its duplicate detection scope will ever be unused at the same time. For example, imagine a situation where threads are constantly releasing and acquiring $n$blocks that prevent the goal $n$block from becoming free. To fix this, we have developed a method called 'hot $n$blocks' where threads altruistically release their $n$block if they are interfering with a better $n$block. We call this enhanced algorithm 'Safe PBNF.'

We use the term 'the *interference scope* of $b$' to refer to the set of $n$blocks that, if acquired, would prevent $b$ from being free. The interference scope includes not only $b$'s successors in the abstract graph, but their predecessors too. In Safe PBNF, whenever a thread checks the heap of free $n$blocks to determine if it should release its current $n$block, it also ensures that its acquired $n$block is better than any of those that it interferes with ($n$blocks whose interference scope the acquired $n$block is in). If it finds a better one, it flags that $n$block as 'hot.' Any thread that finds itself blocking a hot $n$block will release its $n$block in an attempt to free the hot $n$block. For each $n$block $b$ we define $\sigma_h(b)$ to be the number of hot $n$blocks that $b$ is in the interference scope of. If $\sigma_h(b) \neq 0$, $b$ is removed from the heap of free $n$blocks. This ensures that a thread will not acquire an $n$block that is preventing a hot $n$block from becoming free.

Consider, for example, an abstract graph containing four $n$blocks connected in a linear fashion: $A \leftrightarrow B \leftrightarrow C$. A possible execution of PBNF can alternate between a thread expanding from $n$blocks $A$ and $C$. If this situation arrises then $n$blocks $B$ will never be considered free. If only goals are located in $n$block $B$ then, in an infinite search space there may be a livelock. With the "Safe" variant of PBNF, however, when expanding from either $A$ or $C$ a thread will make sure to check the $f$ value of the best open node in $n$block $B$ periodically. If the best node in $B$ is seen to be better than the nodes in $A$ or $C$ then $B$ will be flagged as "hot" and both $n$blocks $A$ and $C$ will no longer be eligable for expansion until after $n$block $B$ has been acquired.

More formally, let $\mathcal{N}$ be the set of all $n$blocks, $Predecessors(x)$ and $Successors(x)$ be the sets of predecessors and successors in the abstract graph of $n$block $x$, $\mathcal{H}$ be the set of all hot $n$blocks, $IntScope(b) = \{l \in \mathcal{N} : \exists x \in Successors(b) : l \in Predecessors(x)\}$ be the interference scope of an $n$block $b$ and $x \prec y$ be a partial order over the $n$blocks where $x \prec y$ iff the minimum $f$ value over all of the open nodes in $x$ is lower than that of $y$. There are three cases to consider when attempting to set an $n$block $b$ to hot with an undirected abstract graph:

1. $\mathcal{H} \cap IntScope(b) = \{\} \wedge \mathcal{H} \cap \{x \in \mathcal{N} : b \in IntScope(x)\} = \{\}$; none of the $n$blocks $b$ interferes with or that interfere with $b$ are hot, so $b$ can be set to hot.

2. $\exists x \in \mathcal{H} : x \in IntScope(b) \wedge x \prec b$; $b$ is interfered with by a better $n$block that is already hot, so $b$ must not be set to hot.

3. $\exists x \in \mathcal{H} : x \in IntScope(b) \wedge b \prec x$; $b$ is interfered with by an $n$block $x$ that is worse than $b$ and $x$ is already hot. $x$ must be un-flagged as hot (updating $\sigma_h$ values appropriately) and in its place $b$ is set to hot.

Directed abstract graphs have two additional cases:

4. $\exists x \in \mathcal{H} : b \in IntScope(x) \wedge b \prec x$; $b$ is interfering with an $n$block $x$ and $b$ is better than $x$ so un-flag $x$ as hot and set $b$ to hot.

5. $\exists x \in \mathcal{H} : b \in IntScope(x) \wedge x \prec b$; $b$ is interfering with an $n$block $x$ and $x$ is better than $b$ so do not set $b$ to hot.

This scheme ensures that there are never two hot $n$blocks interfering with one another and that the $n$block that is set to hot is the best $n$block in its interference scope. As we verify below, this approach guarantees the property that if an $n$block is flagged as hot it will eventually become free. Full pseudo-code for Safe PBNF is given in Appendix A.

### 3.2 Correctness of PBNF

Given the complexity of parallel shared-memory algorithms, it can be reassuring to have proofs of correctness. In this subsection we will verify that PBNF exhibits various desirable properties:

#### 3.2.1 SOUNDNESS

Soundness holds trivially because no solution is returned that does not pass the goal test.

#### 3.2.2 DEADLOCK

There is only one lock in PBNF and the thread that currently holds it never attempts to acquire it a second time, so deadlock cannot arise.

#### 3.2.3 LIVELOCK

Because the interaction between the different threads of PBNF can be quite complex, we modeled the system using the TLA$^+$ (Lamport, 2002) specification language. Using the TLC model checker (Yu, Manolios, & Lamport, 1999) we were able to demonstrate a sequence of states that can give rise to a livelock in plain PBNF. Using a similar model we were unable to find an example of livelock in Safe PBNF when using up to three threads and 12 $n$blocks in an undirected ring-shaped abstract graph and up to three threads and eight $n$blocks in a directed graph.

In our model the state of the system is represented with four variables: *state*, *acquired*, *isHot* and *Succs*. The *state* variable contains the current action that each thread is performing (either *search* or *nextblock*). The *acquired* variable is a function from each thread to the ID of its acquired $n$block or the value *None* if it currently does not have an $n$block. The variable *isHot* is a function from $n$blocks to either TRUE or FALSE depending on whether or not the given $n$block is flagged as hot. Finally, the *Succs* variable gives the set of successor $n$blocks for each $n$block in order to build the $n$block graph.

The model has two actions: *doSearch* and *doNextBlock*. The *doSearch* action models the search stage performed by a PBNF thread. Since we were interested in determining if there is a livelock, this action abstracts away most of the search procedure and merely models that the thread may choose a valid $n$block to flag as hot. After setting an $n$block to hot, the thread changes its state so that the next time it is selected to perform an action it will try to acquire a new $n$block. The *doNextBlock* simulates a thread choosing its next $n$block if there is one available. After a thread acquires an $n$block (if one was free) it sets its state so that the next time it is selected to perform an action it will search.

The TLA$^+$ source of the model is located in Appendix B.

**Formal proof**: In addition to model checking, the TLA$^+$ specification language is designed to allow for formal proofs of properties. This allows properties to be proved for an unbounded space. Using our model we have completed a formal proof that a hot $n$block will eventually become free regardless of the number of threads or the abstract graph. We present here an English summary. First, we need a helpful lemma:

**Lemma 1** *If an nblock $n$ is hot, there is at least one other nblock in its interference scope that is in use. Also, $n$ is not interfering with any other hot nblocks.*

**Proof:** Initially no $n$blocks are hot. This can change only while a thread searches or when it releases an $n$block. During a search, a thread can only set $n$ to hot if it has acquired an $n$block $m$ that is in the interference scope of $n$. Additionally, a thread may only set $n$ to hot if it does not create any interference with another hot $n$block. During a release, if $n$ is hot, either the final acquired $n$block in its interference scope is released and $n$ is no longer hot, or $n$ still has at least one busy $n$block in its interference scope. □

Now we are ready for the key theorem:

**Theorem 1** *If an nblock $n$ becomes hot, it will eventually be added to the free list and will no longer be hot.*

**Proof:** We will show that the number of acquired $n$blocks in the interference scope of a hot $n$block $n$ is strictly decreasing. Therefore, $n$ will eventually become free.

Assume an $n$block $n$ is hot. By Lemma 1, there is a thread $p$ that has an $n$block in the interference scope of $n$, and $n$ is not interfering with or interfered by any other hot $n$blocks. Assume that a thread $q$ does not have an $n$block in the interference scope of $n$. There are four cases:

1. $p$ searches its $n$block. $p$ does not acquire a new $n$block and therefore the number of $n$blocks preventing $n$ from becoming free does not increase. If $p$ sets an $n$block $m$ to hot, $m$ is not in the interference scope of $n$ by Lemma 1. $p$ will release its $n$block after it sees that $n$ is hot (see case 2).

2. $p$ releases its $n$block and acquires a new $n$block $m$ from the free list. The number of acquired $n$blocks in the interference scope of $n$ decreases by one as $p$ releases its $n$block. Since $m$, the new $n$block acquired by $p$, was on the free list, it is not in the interference scope of $n$.

3. $q$ searches its $n$block. $q$ does not acquire a new $n$block and therefore the number of $n$blocks preventing $n$ from becoming free does not increase. If $q$ sets an $n$block $m$ to hot, $m$ is not in the interference scope of $n$ by Lemma 1.

4. $q$ releases its $n$block (if it had one) and acquires a new $n$block $m$ from the free list. Since $m$, the new $n$block acquired by $q$, was on the free list, it is not in the interference scope of $n$ and the number of $n$blocks preventing $n$ from becoming free does not increase.

□

We can now prove the progress property that we really care about:

**Theorem 2** *A node $n$ with minimum $f$ value will eventually be expanded.*

**Proof:** We consider $n$'s $n$block. There are three cases:

1. The $n$block is being expanded. Because $n$ has minimum $f$, it will be at the front of *open* and will be expanded.

2. The $n$block is free. Because it holds the node with minimum $f$ value, it will be at the front of the free list and selected next for expansion, reducing to case 1.

3. The $n$block is not on the free list because it is in the interference scope of another $n$block that is currently being expanded. When the thread expanding that $n$block checks its interference scope, it will mark the better $n$block as hot. By Theorem 1, we will eventually reach case 2.

<div align="right">□</div>

### 3.2.4 COMPLETENESS

This follows easily from liveness:

**Corollary 1** *If the heuristic is admissible or the search space is finite, a goal will be returned if one is reachable.*

**Proof:** If the heuristic is admissible, we inherit the completeness of serial A* (Nilsson, 1980) by Theorem 2. Nodes are only re-expanded if their $g$ value has improved, and this can happen only a finite number of times, so a finite number of expansions will suffice to exhaust the search space. □

### 3.2.5 OPTIMALITY

Because PBNF's expansion order is not strictly best-first, it operates like an anytime algorithm, and its optimality follows the same argument as that for algorithms such as Anytime A* (Hansen & Zhou, 2007).

**Theorem 3** *PBNF will only return optimal solutions.*

**Proof:** After finding an incumbent solution, the search continues to expand nodes until the minimum $f$ value among all frontier nodes is greater than or equal to the incumbent solution cost. This means that the search will only terminate with the optimal solution. □

Before discussing how to adapt PBNF to suboptimal and anytime search, we first evaluate its performance on optimal problem solving.

## 4. Empirical Evaluation: Optimal Search

We have implemented and tested the parallel heuristic search algorithms described above on three different benchmark domains: grid pathfinding, the sliding tile puzzle, and STRIPS planning. We will discuss each domain in turn. With the exception of the planning domain, the algorithms were programmed in C++ using the POSIX threading library and run on dual quad-core Intel Xeon E5320 1.86GHz processors with 16Gb RAM. For the planning results the algorithms were written independently in C from the pseudo code in Appendix A. This gives us additional confidence in the correctness of the pseudo code and our performance claims. The planning experiments were run on dual quad-core Intel Xeon X5450 3.0GHz processors limited to roughly 2GB of RAM. All open lists and free lists were implemented as binary heaps except in PSDD and IDPSDD which used a queue giving them less overhead since they do not require access to minimum valued elements. All closed lists were implemented as hash tables. PRA* and APRA* used queues for incoming nodes, and a hash table was used to detect duplicates in both open and closed. For grids and sliding tiles, we used the jemalloc library (Evans, 2006), a special multi-thread-aware malloc implementation, instead of the standard glibc (version 2.7) malloc, because we found that the latter scales poorly above 6 threads. We configured jemalloc to use 32 memory arenas per CPU. In planning, a custom memory manager was used which is also thread-aware and uses a memory pool for each thread.

On grids and sliding tiles abstractions were hand-coded and, $n$block data structures were created lazily, so only the visited part of abstract graph was instantiated. The time taken to create the abstraction is accounted for in all of the wall time measurements for these two domains. In STRIPS planning the abstractions were created automatically and the creation times for the abstractions are reported separately as described in Section 4.5.

## 4.1 Tuning PBNF

In this section we present results for a set of experiments that we designed to test the behavior of PBNF as some of its parameters are changed. We study the effects of the two important parameters of the PBNF algorithm: minimum expansions required before switching to search a new $n$block and the size of the abstraction. This study used twenty 5000x5000 four-connected grid pathfinding instances with unit cost moves where each cell has a 0.35 probability of being an obstacle. The heuristic used was the Manhattan distance to the goal location. Error bars in the plots show 95% confidence intervals and the legends are sorted by the mean of the dependent variable in each plot.

In the PBNF algorithm, each thread must perform a minimum number of expansions before it is able to acquire a new $n$block for searching. Requiring more expansions between switches is expected to reduce the contention on the $n$block graph's lock but could increase the total number of expanded nodes. We created an instrumented version of the PBNF algorithm that tracks the time that the threads have spent trying to acquire the lock and the amount of time that threads have spent waiting for a free $n$block. We fixed the size of the abstraction to 62,500 $n$blocks and varied the number of threads (from 1 to 8) and minimum expansions (1, 8, 16, 32 and 64 minimum expansions).

The upper left panel in Figure 4 shows the average amount of *CPU time* in seconds that each thread spent waiting to acquire the lock (y-axis) as the minimum expansions parameter was increased (x-axis). Each line in this plot represents a different number of threads. We can see that the configuration which used the most amount of time trying to acquire the lock was with eight threads and one minimum expansion. As the number of threads decreased, there was less contention on the lock as there were fewer threads to take it. As the number of minimum required expansions increased the contention was also reduced. Around eight minimum expansions the benefit of increasing the value further seemed to greatly diminish.

The upper right panel of Figure 4 shows the results for the CPU time spent waiting for a free $n$block (y-axis) as minimum expansions was increased (x-axis). This is different than the amount of time waiting on the lock because, in this case, the thread successfully acquired the lock but then found that there were no free $n$blocks available to search. We can see that the configuration with eight threads and one for minimum expansions caused the longest amount of time waiting for a free $n$block. As the number of threads decreased and as the required number of minimum expansions increased the wait time decreased. The amount of time spent waiting, however, seems fairly insignificant because it is an order of magnitude smaller than the lock time. Again, we see that around eight minimum expansions the benefit of increasing seemed to diminish.

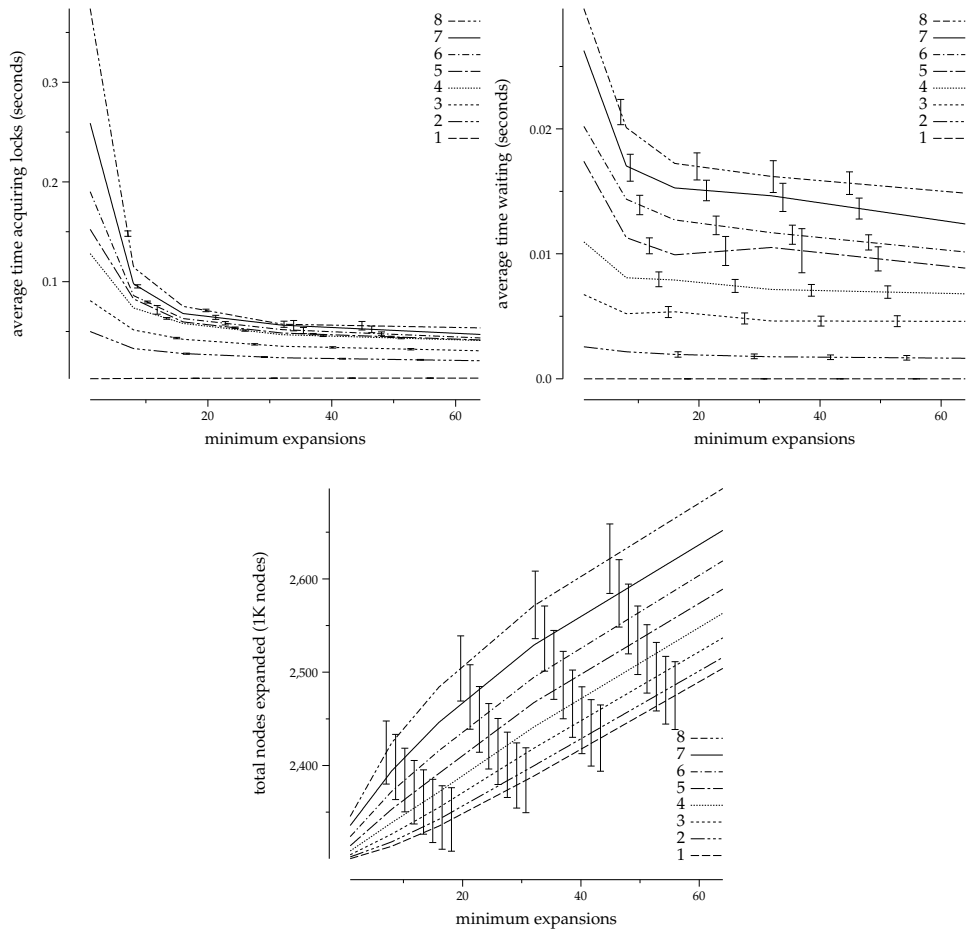

Figure 4: PBNF locking behavior vs minimum expansions on grid pathfinding with 62,500 $n$blocks. Each line represents a different number of threads.

The final panel, on the bottom in Figure 4, shows the total number of nodes expanded (y-axis, which is in thousands of nodes) as minimum expansions was increased. Increasing the minimum number of expansions that a thread must make before switching to an $n$block with better nodes caused the search algorithm to explore more of the space that may not have been covered by a strict best-first search. As more of these "speculative" expansions were performed the total number of nodes encountered during the search increased. We can also see that adding threads increased the number of expanded nodes too.

From the results of this experiment it appears that requiring more than eight expansions before switching $n$blocks had a decreasing benefit with respect to locking and waiting time. In our non-instrumented implementation of PBNF we found that slightly greater values for the minimum expansion parameter lead to the best total wall times. For each domain below we use the value that gave the best total wall time in the non-instrumented PBNF implementation.

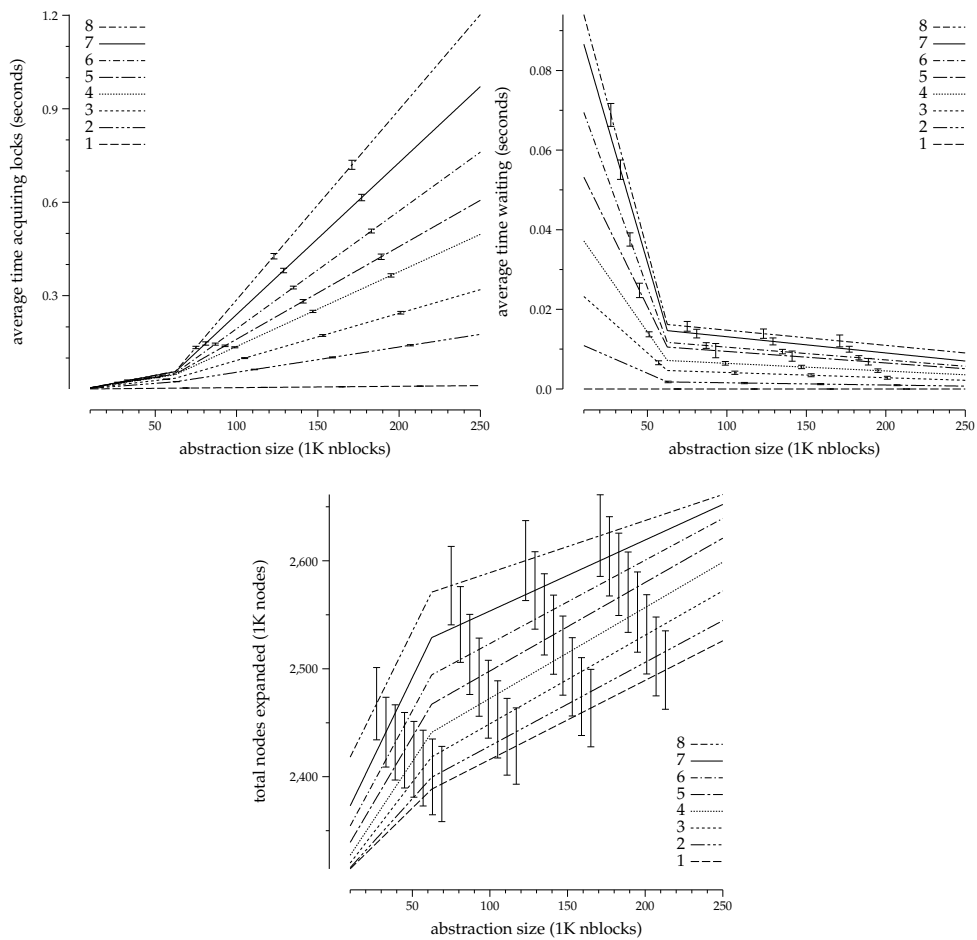

Figure 5: PBNF abstraction size: 5000x5000 grid pathfinding, 32 minimum expansions.

Since PBNF uses abstraction to decompose a search space it is also important to understand the effect of abstraction size on search performance. Our hypothesis was that using too few abstract states would lead to only a small number of free $n$blocks therefore making threads spend a lot of time waiting for an $n$block to become free. On the other hand, if there are too many abstract states then there will be too few nodes in each $n$block. If this happens, threads will perform only a small amount of work before exhausting the open nodes in their $n$block and being forced to switch to a new portion of the search space. Each time a thread must switch $n$blocks the contention on the lock is increased. Figure 5 shows the results of an experiment that was performed to verify this theory. In each plot we have fixed the minimum expansions parameter to 32 (which gave the best total wall time on grid pathfinding) and varied the number of threads (from 1 to 8) and the size of the abstraction (10,000, 62,500 and 250,000 $n$blocks).

The upper left panel of Figure 5 shows a plot of the amount of CPU seconds spent trying to acquire the lock (y-axis) versus the size of the abstraction (x-axis). As expected, when the abstraction was very coarse there was little time spent waiting on the lock, but as the size of the abstraction grew and the number of threads increased the amount of time spent locking increased. At eight threads with 250,000 $n$blocks over 1 second of CPU time was spent waiting to acquire the lock. We suspect that this is because threads were exhausting all open nodes in their $n$blocks and were, therefore, being forced to take the lock to acquire a new portion of the search space.

The upper right panel of Figure 5 shows the amount of time that threads spent waiting for an $n$block to become free after having successfully acquired the lock only to find that no $n$blocks are available. Again, as we suspected, the amount of time that threads wait for a free $n$block decreases as the abstraction size is increased. The more available $n$blocks, the more disjoint portions of the search space will be available. As with our experiments for minimum expansions, the amount of time spent waiting seems to be relatively insignificant compared to the time spent acquiring locks.

The bottom panel in Figure 5 shows that the number of nodes that were expanded increased as the size of the abstraction was increased. For finer grained abstractions the algorithm expanded more nodes. This is because each time a thread switches to a new $n$block it is forced to perform at least the minimum number of expansions, therefore the more switches, the more forced expansions.

## 4.2 Tuning PRA*

We now turn to looking at the performance impact on PRA* of abstraction and asynchronous communication. First, we compare PRA* with and without asynchronous communication. Results from a set of experiments on twenty 5000x5000 grid pathfinding and a set of 250 random 15-puzzle instances that were solvable by A* in 3 million expansions are shown in Figure 6. The line labeled *sync. (PRA\*)* used synchronous communication, *async. sends*, used synchronous receives and asynchronous sends, *async. receives*, used synchronous sends and asynchronous receives and *async. (HDA\*)*, used asynchronous communication for both sends and receives. As before, the legend is sorted by the mean performance and the error bars represent the 95% confidence intervals on the mean. The vertical lines in the plots for the life cost grid pathfinding domains show that these configurations were unable to solve instances within the 180 second time limit.

The combination of both asynchronous sends and receives provided the best performance. We can also see from these plots that making sends asynchronous provided more of a benefit than making receives asynchronous. This is because, without asynchronous sends, each node that is generated will stop the generating thread in order to communicate. Even if communication is batched, each send may be required to go to a separate neighbor and therefore a single send operation may be required per-generation. For receives, the worst case is that the receiving thread must stop at each expansion to receive the next batch nodes. Since the branching factor in a typical search space is approximately constant there will be approximately a constant factor more send communications as there are receive communications in the worst case. Therefore, making sends asynchronous reduces the communication cost more than receives.

Figure 7 shows the results of an experiment that compares PRA* using abstraction to distribute nodes among the threads versus PRA* with asynchronous communication. The lines are labeled as follows: *sync. (PRA\*)* used only synchronous communication, *async. (HDA\*)* used only asynchronous communication and *sync. with abst. (APRA\*)* used only synchronous communication and used abstraction to distribute nodes among the threads and *async. and abst. (AHDA\*)* used a combination of asynchronous communication and abstraction. Again, the vertical lines in the plots for the life cost grid pathfinding domains show that these configurations were unable to solve instances within the 180 second time limit.

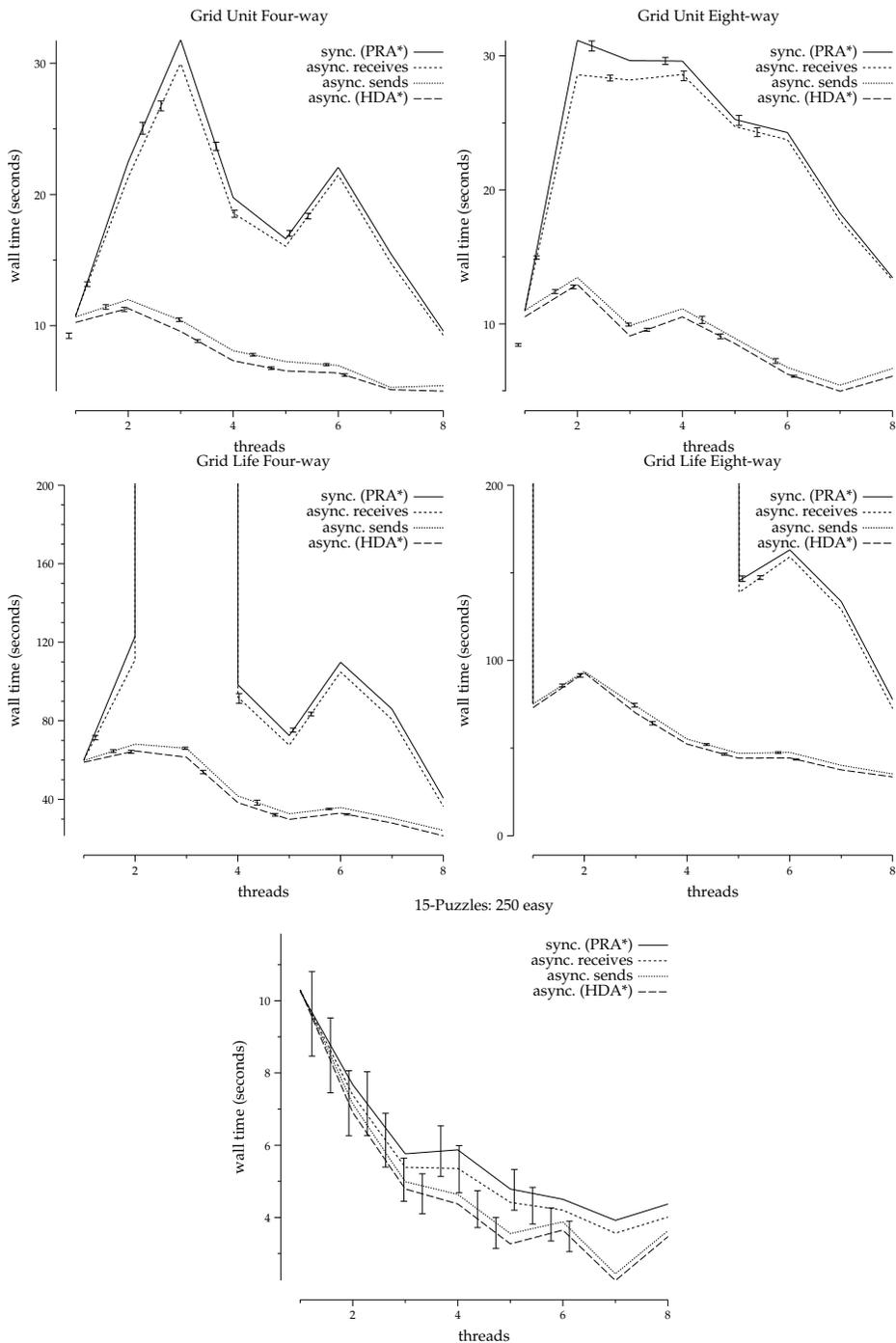

Figure 6: PRA* synchronization: 5000x5000 grids and easy sliding tile instances.

It is clear from these plots that the configurations of PRA* that used abstraction gave better performance than PRA* without abstraction in the grid pathfinding domain. The reason for this is because the abstraction in grid pathfinding will often assign successors of a node being expanded back to the thread that generated them. When this happens no communication is required and the nodes can simply be checked against the local closed list and placed on the local open list if they are not duplicates. With abstraction, the only time that communication will be required is when a node on the "edge" of an abstract state is expanded. In this case, *some* of the children will map into a different abstract state and communication will be required. This experiment also shows that the benefits of abstraction were greater than the benefits of asynchronous communication in the grid pathfinding problems. We see the same trends on the sliding tile instances, however they are not quite as pronounced; the confidence intervals often overlap.

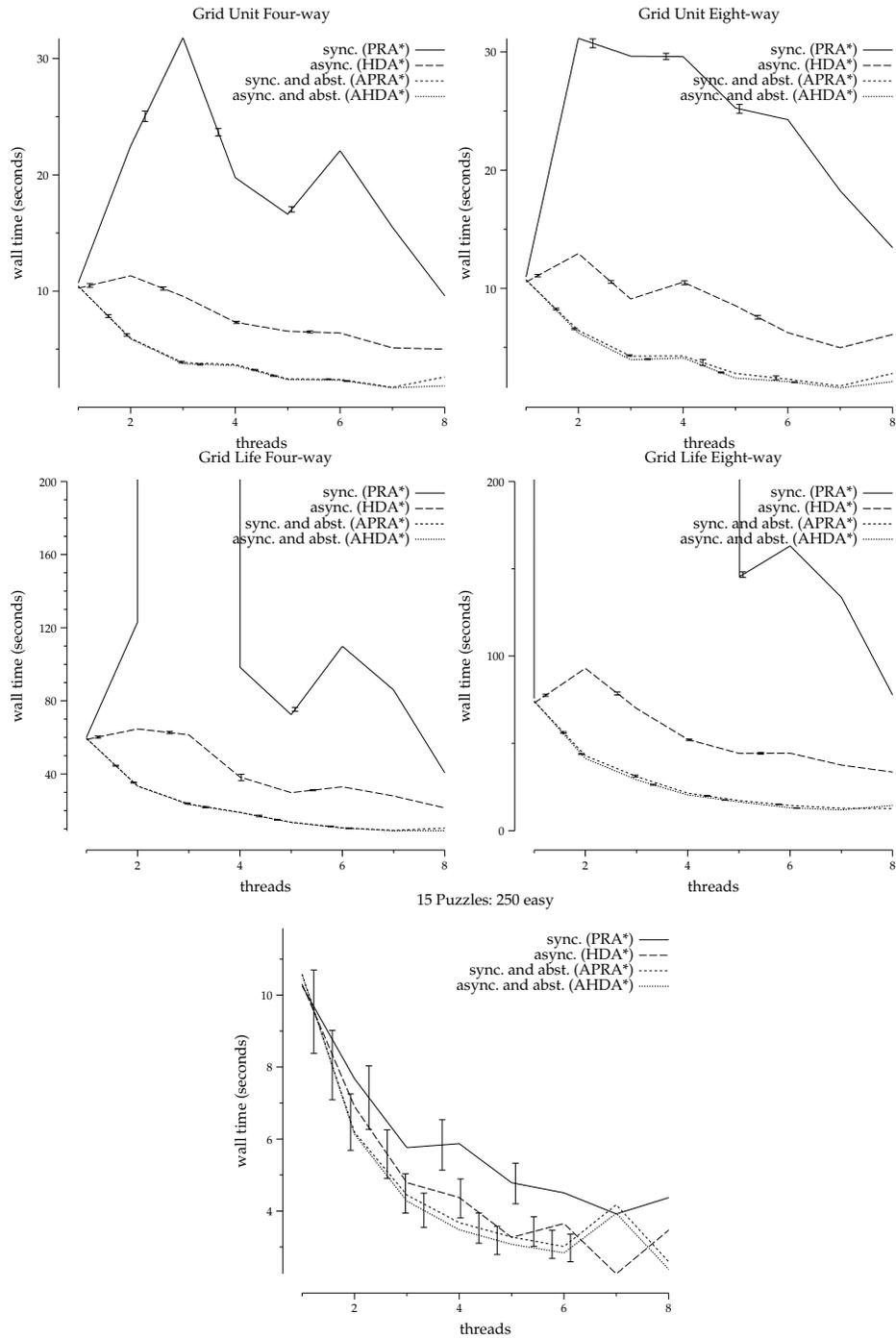

Figure 7: PRA* abstraction: 5000x5000 grids and easy sliding tile instances.

Overall, it appears that the combination of PRA* with both abstraction for distributing nodes among the different threads and using asynchronous communication gave the best performance. In the following section we show the results of a comparison between this variant of PRA*, the Safe PBNF algorithm and the best-first variant of PSDD.

## 4.3 Grid Pathfinding

In this section, we evaluate the parallel algorithms on the grid pathfinding domain. The goal of this domain is to navigate through a grid from an initial location to a goal location while avoiding obstacles. We used two cost models (discussed below) and both four-way and eight-way movement. On the four-way grids, cells were blocked with a probability of 0.35 and on the eight-way grids cells were blocked with a probability of 0.45. The abstraction function that was used maps blocks of adjacent cells to the same abstract state, forming a coarser abstract grid overlaid on the original space. The heuristic was the Manhattan distance to the goal location. The hash values for states (which are used to distribute nodes in PRA* and HDA*) are computed as: $x \cdot y_{max} + y$ of the state location. This gives a minimum perfect hash value for each state. For this domain we were able to tune the size of the abstraction and our results show execution with the best abstraction size for each algorithm where it is relevant.

### 4.3.1 FOUR-WAY UNIT COST

In the unit-cost model, each move has the same cost: one.

**Less Promising Algorithms**    Figure 8, shows a performance comparison between algorithms that, on average, were slower than serial A*. These algorithms were tested on 20 unit-cost four-way movement 1200x2000 grids with the start location in the bottom left corner and the goal location in the bottom right. The x-axis shows the number of threads used to solve each instance and the y-axis shows the mean wall clock time in seconds. The error bars give a 95% confidence interval on the mean wall clock time and the legend is sorted by the mean performance.

From this figure we can see that PSDD gave the worst average solution times. We suspect that this was because the lack of a tight upper bound which PSDD uses for pruning. We see that A* with a shared lock-free open and closed list (LPA*) took, on average, the second longest amount of time to solve these problems. LPA*'s performance improved up to 5 threads and then started to drop off as more threads were added. The overhead of the special lock-free memory manager along with the fact that access to the lock-free data structures may require back-offs and retries could account for the poor performance compared to serial A*. The next algorithm, going down from the top in the legend, is KBFS which slowly increased in performance as more threads were added however it was not able to beat serial A*. A simple parallel A* implementation (PA*) using locks on the open and closed lists performed worse as threads were added until about four where it started to give a very slow performance increase matching that of KBFS. The PRA* algorithm using a simple state representation based hashing function gave the best performance in this graph but it was fairly erratic as the number of threads changed, sometimes increasing and sometimes decreasing. At 6 and 8 threads, PRA* was faster than serial A*.

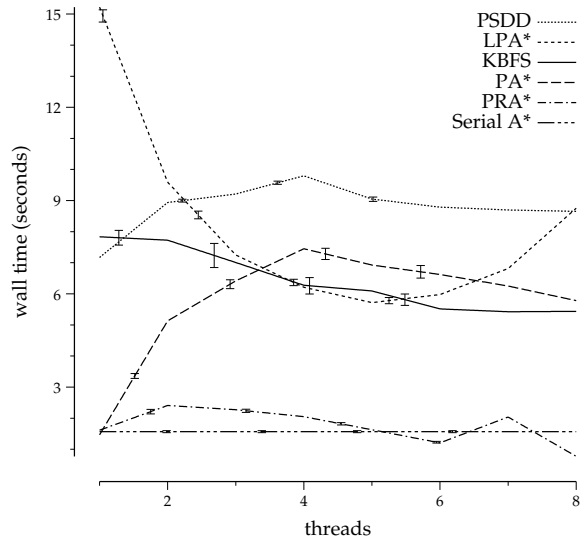

Figure 8: Simple parallel algorithms on unit cost, four-way 2000x1200 grid pathfinding.

We have also implemented the IDPSDD algorithm which tries to find the upper bound for a PSDD search using iterative deepening, but the results are not shown on the grid pathfinding domains. The non-geometric growth in the number of states when increasing the cost bound leads to very poor performance with iterative deepening on grid pathfinding. Due to the poor performance of the above algorithms, we do not show their results in the remaining grid, tiles or planning domains (with the exception of PSDD which makes a reappearance in the STRIPS planning evaluation of Section 4.5, where we supply it with an upper bound).

**More Promising Algorithms** The upper left plot in Figure 9 shows the performance of algorithms on unit-cost four-way grid pathfinding problems. The y-axis represents the speedup over serial A* and the x-axis shows the number of threads in use for each data point. Error bars indicate 95% confidence intervals on the mean over 20 different instances. Algorithms in the legend are ordered by their average performance. The line labeled "Perfect speedup" shows a perfect linear speedup where each additional thread increases the performance linearly.

A more practical reference point for speedup is shown by the "Achievable speedup" line. On a perfect machine with $n$ processors, running with $n$ cores should take time that decreases linearly with $n$. On a real machine, however, there are hardware considerations such as memory bus contention that prevent this $n$-fold speedup. To estimate this overhead for our machines, we ran sets of $n$ independent A* searches in parallel for $1 \leq n \leq 8$ and calculated the total time for each set to finish. On a perfect machine all of these sets would take the same time as the set with $n = 1$. We compute the "Achievable speedup" with the ratio of the actual completion times to the time for the set with $n = 1$. At $t$ threads given the completion times for the sets, $\langle C_1, C_2, ..., C_n \rangle$, $achievable\_speedup(t) = \frac{t \cdot C_1}{C_t}$.

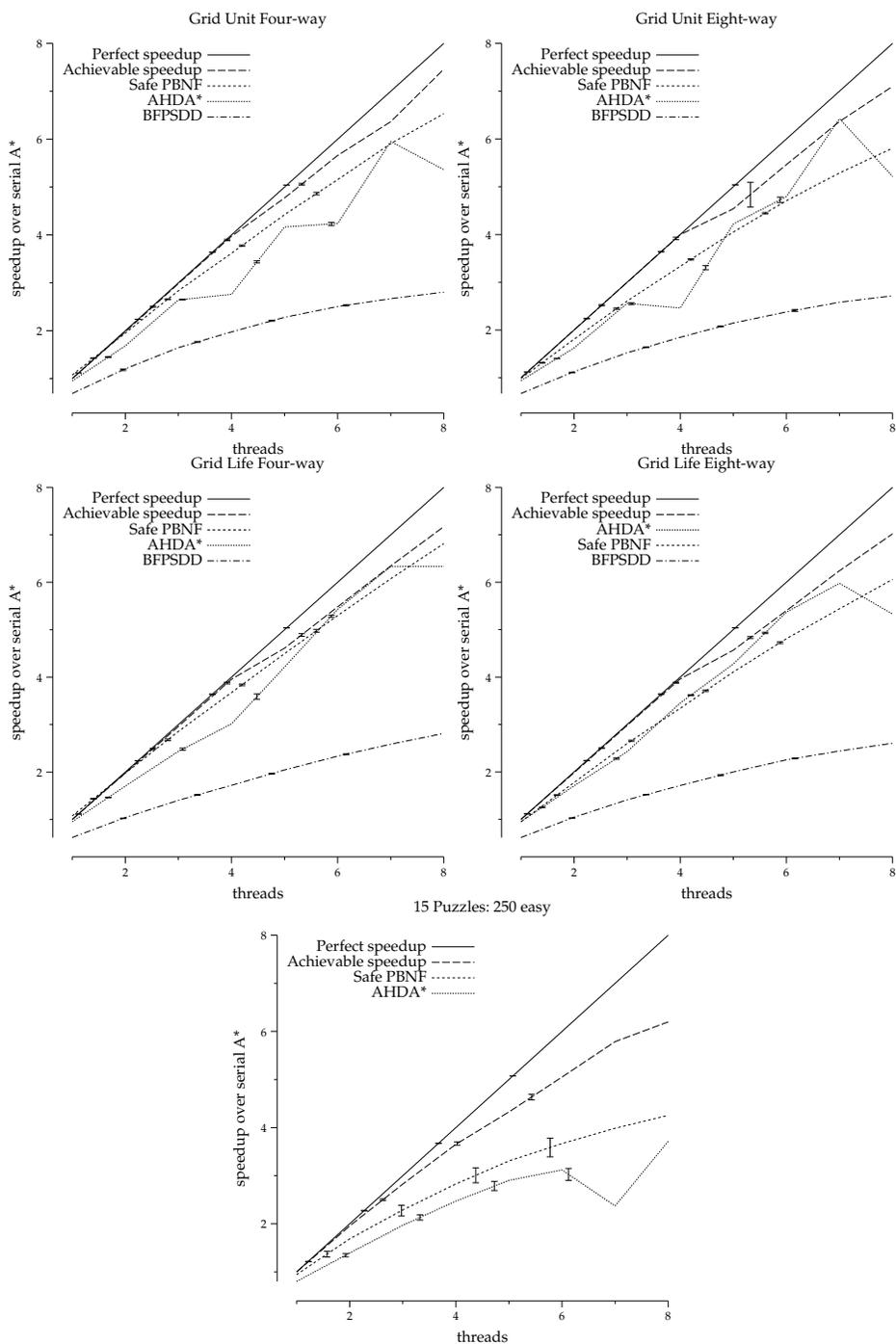

Figure 9: Speedup results on grid pathfinding and the sliding tile puzzle.

The upper left panel shows a comparison between AHDA* (PRA* with asynchronous communication and abstraction), BFPSDD and Safe PBNF algorithm on the larger (5000x5000) unit-cost four-way problems. Safe PBNF was superior to any of the other algorithms, with steadily decreasing solution times as threads were added and an average speedup over serial A* of more than 6x when using eight threads. AHDA* had less stable performance, sometimes giving a sharp speedup increase and sometimes giving a decreased performance as more threads were added. At seven threads where AHDA* gave its best performance, it was able to reach 6x speedup over serial A* search. The BFPSDD algorithm solved problems faster as more threads were added however it was not as competitive as PBNF and AHDA* giving no more than 3x speedup over serial A* with eight threads.

### 4.3.2 FOUR-WAY LIFE COST

Moves in the life cost model have a cost of the row number of the state where the move was performed—moves at the top of the grid are free, moves at the bottom cost 4999 (Ruml & Do, 2007). This differentiates between the shortest and cheapest paths which has been shown to be a very important distinction (Richter & Westphal, 2010; Cushing, Benton, & Kambhampati, 2010). The left center plot in Figure 9 shows these results in the same format as for the unit-cost variant – number of threads on the x axis and speedup over serial A* on the y axis. On average, Safe PBNF gave better speedup than AHDA*, however AHDA* outperformed PBNF at six and seven threads. At eight threads, however, APRA* did not perform better than at seven threads. Both of these algorithms achieve speedups that are very close to the "Achievable speedup" for this domain. Again BFPSDD gave the worst performance increase as more threads were added reaching just under 3x speedup.

### 4.3.3 EIGHT-WAY UNIT COST

In eight-way movement path planning problems, horizontal and vertical moves have cost 1, but diagonal movements cost $\sqrt{2}$. These real-valued costs make the domain different from the previous two path planning domains. The upper right panel of Figure 9 shows number of threads on the x axis and speedup over serial A* on the y axis for the unit cost eight-way movement domain. We see that Safe PBNF gave the best average performance reaching just under 6x speedup at eight threads. AHDA* did not outperform Safe PBNF on average, however it was able to achieve a just over 6x speedup over serial A* at seven threads. Again however, we see that AHDA* did not give very stable performance increases with more threads. BFPSDD improved as threads were added out to eight but it never reached more than 3x speedup.

### 4.3.4 EIGHT-WAY LIFE COST

This model combines the eight-way movement and the life cost models; it tends to be the most difficult path planning domain presented in this paper. The right center panel of Figure 9 shows threads on the x axis and speedup over serial A* on the y axis. AHDA* gave the best average speedup over serial A* search, peaking just under 6x speedup at seven threads. Although it outperformed Safe PBNF on average at eight threads AHDA* has a sharp decrease in performance reaching down to almost 5x speedup where Safe PBNF had around 6x speedup over serial A*. BFPSDD again peaks at just under 3x speedup at eight threads.

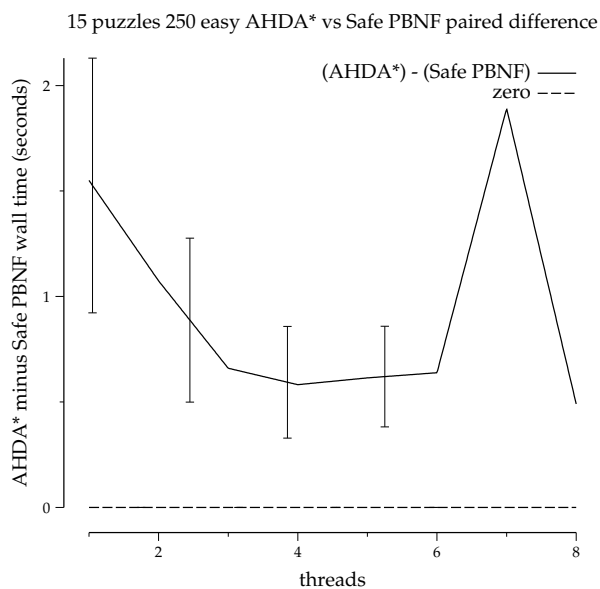

15 puzzles 250 easy AHDA* vs Safe PBNF paired difference

Figure 10: Comparison of wall clock time for Safe PBNF versus AHDA* on the sliding tile puzzle.

## 4.4 Sliding Tile Puzzle

The sliding tile puzzle is a common domain for benchmarking heuristic search algorithms. For these results, we use 250 randomly generated 15-puzzles that serial A* was able to solve within 3 million expansions.

The abstraction used for the sliding tile puzzles ignores the numbers on a set of tiles. For example, the results shown for Safe PBNF in the bottom panel of Figure 9 use an abstraction that looks at the position of the blank, one and two tiles. This abstraction gives 3360 $n$blocks. In order for AHDA* to get the maximum amount of expansions that map back to the expanding thread (as described above for grids), its abstraction uses the one, two and three tile. Since the position of the blank is ignored, any state generation that does not move the one, two or three tiles will generate a child into the same $n$block as the parent therefore requiring no communication. The heuristic that was used in all algorithms was the Manhattan distance heuristic. The hash value used for tiles states was a perfect hash value based on the techniques presented by Korf and Schultze (2005).

The bottom panel of Figure 9 shows the results for AHDA*, and Safe PBNF on these sliding tiles puzzle instances. The plot has the number of threads on the x axis and the speedup over serial A* on the y axis. Safe PBNF had the best mean performance but there was overlap in the confidence intervals with AHDA*. BFPSDD was unable to show a speedup over serial A* and its performance was not shown in this plot.

Because sliding tile puzzles vary so much in difficulty, in this domain we also did a paired-difference test, shown in Figure 10. The data used for Figure 10 was collected on the same set of runs as shown in the bottom panel of Figure 9. The y-axis in this figure, however, is the average, over all instances, of the time that AHDA* took on that instance minus the time that Safe PBNF took. This paired test gives a more powerful view of the algorithms' relative performance. Values greater than 0.0 represent instances where Safe PBNF was faster than AHDA* and values lower than 0.0 represent those instances where AHDA* was faster. The error bars show the 95% confidence interval on the mean. We can clearly see that the Safe PBNF algorithm was significantly faster than AHDA* across all numbers of threads from 1 to 8.

## 4.5 STRIPS Planning

In addition to the path planning and sliding tiles domains, the algorithms were also embedded into a domain-independent optimal sequential STRIPS planner. In contrast to the previous two domains where node expansion is very quick and therefore it is difficult to achieve good parallel speedup, node expansion in STRIPS planning is relatively slow. The planner used in these experiments uses regression and the max-pair admissible heuristic of Haslum and Geffner (2000). The abstraction function used in this domain is generated dynamically on a per-problem basis and, following Zhou and Hansen (2007), this time was not taken into account in the solution times presented for these algorithms. The abstraction function is generated by greedily searching in the space of all possible abstraction functions (Zhou & Hansen, 2006). Because the algorithm needs to evaluate one candidate abstraction for each of the unselected state variables, it can be trivially parallelized by having multiple threads work on different candidate abstractions.

Table 1 presents the results for A*, AHDA*, PBNF, Safe PBNF, PSDD (given an optimal upper bound for pruning and using divide-and-conquer solution reconstruction), APRA* and BFPSDD. The values of each cell are the total wall time in seconds taken to solve each instance. A value of 'M' indicates that the program ran out of memory. The best result on each problem and results within 10% of the best are marked in **bold**. Generally, all of the parallel algorithms were able to solve the instances faster as they were allowed more threads. All of the parallel algorithms were able to solve instances much faster than serial A* at seven threads. The PBNF algorithm (either PBNF or Safe PBNF) gave the best solution times in all but three domains. Interestingly, while plain PBNF was often a little faster than the safe version, it failed to solve two of the problems. This is most likely due to livelock, although it could also simply be because the hot $n$blocks fix forces Safe PNBF to follow a different search order than PBNF. AHDA* tended to give the second-best solution times, followed by PSDD which was given the optimal solution cost up-front for pruning. BFPSDD was often better than APRA*,

The column, labeled "Abst." shows the time that was taken by the parallel algorithms to serially generate the abstraction function. Even with the abstraction generation time added on to the solution times all of the parallel algorithms outperform A* at seven threads, except in the block-14 domain where the time taken to generate the abstraction actually was longer than the time A* took to solve the problem.

## 4.6 Understanding Search Performance

We have seen that the PBNF algorithm tends to have better performance than the AHDA* algorithm for optimal search. In this section we show the results of a set of experiments that attempts to determine which factors allow PBNF to perform better in these domains. We considered three hypotheses. First, PBNF may achieve better performance because it expands fewer nodes with $f$ values greater than the optimal solution cost. Second, PBNF may achieve better search performance because it tends to have many fewer nodes on each priority queue than AHDA*. Finally, PBNF may achieve better search performance because it spends less time coordinating between threads. In the following subsections we show the results of experiments that we performed to test our

713

| threads | A* 1 | AHDA* 1 | 3 | 5 | 7 | PBNF 1 | 3 | 5 | 7 |
|---|---|---|---|---|---|---|---|---|---|
| logistics-6 | 2.30 | 1.44 | 0.70 | 0.48 | **0.40** | 1.27 | 0.72 | 0.58 | 0.53 |
| blocks-14 | 5.19 | 7.13 | 5.07 | 2.25 | **2.13** | 6.28 | 3.76 | 2.70 | 2.63 |
| gripper-7 | 117.78 | 59.51 | 33.95 | 15.97 | 12.69 | 39.66 | 16.43 | 10.92 | **8.57** |
| satellite-6 | 130.85 | 95.50 | 33.59 | 24.11 | 18.24 | 68.14 | 34.15 | 20.84 | 16.57 |
| elevator-12 | 335.74 | 206.16 | 96.82 | 67.68 | 57.10 | 156.64 | 56.25 | 34.84 | **26.72** |
| freecell-3 | 199.06 | 147.96 | 93.55 | 38.24 | **27.37** | 185.68 | 64.06 | 44.05 | 36.08 |
| depots-7 | M | 299.66 | 126.34 | 50.97 | 39.10 | M | M | M | M |
| driverlog-11 | M | 315.51 | 85.17 | 51.28 | 48.91 | M | M | M | M |
| gripper-8 | M | 532.51 | 239.22 | 97.61 | 76.34 | 229.88 | 95.63 | 60.87 | **48.32** |

| threads | SafePBNF 1 | 3 | 5 | 7 | PSDD 1 | 3 | 5 | 7 |
|---|---|---|---|---|---|---|---|---|
| logistics-6 | 1.17 | 0.64 | 0.56 | 0.62 | 1.20 | 0.78 | 0.68 | 0.64 |
| blocks-14 | 6.21 | 2.69 | **2.20** | **2.02** | 6.36 | 3.57 | 2.96 | 2.87 |
| gripper-7 | 39.58 | 16.87 | 11.23 | **9.21** | 65.74 | 29.37 | 21.88 | 19.19 |
| satellite-6 | 77.02 | 24.09 | 17.29 | **13.67** | 61.53 | 23.56 | 16.71 | **13.26** |
| elevator-12 | 150.39 | 53.45 | 34.23 | **27.02** | 162.76 | 62.68 | 43.34 | 36.66 |
| freecell-3 | 127.07 | 47.10 | 38.07 | 37.02 | 126.31 | 53.76 | 45.47 | 43.71 |
| depots-7 | 156.36 | 63.04 | 42.91 | **34.66** | 159.98 | 73.00 | 57.65 | 54.70 |
| driverlog-11 | 154.15 | 59.98 | 38.84 | **31.22** | 155.93 | 63.20 | 41.85 | **34.02** |
| gripper-8 | 235.46 | 98.21 | 63.65 | **51.50** | 387.81 | 172.01 | 120.79 | 105.54 |

| threads | APRA* 1 | 3 | 5 | 7 | BFPSDD 1 | 3 | 5 | 7 | Abst. 1 |
|---|---|---|---|---|---|---|---|---|---|
| logistics-6 | 1.44 | 0.75 | 1.09 | 0.81 | 2.11 | 1.06 | 0.79 | 0.71 | 0.42 |
| blocks-14 | 7.37 | 5.30 | 3.26 | 2.92 | 7.78 | 4.32 | 3.87 | 3.40 | 7.9 |
| gripper-7 | 62.61 | 43.13 | 37.62 | 26.78 | 41.56 | 18.02 | 12.21 | 10.20 | 0.8 |
| satellite-6 | 95.11 | 42.85 | 67.38 | 52.82 | 62.01 | 24.06 | 20.43 | **13.54** | 1 |
| elevator-12 | 215.19 | 243.24 | 211.45 | 169.92 | 151.50 | 58.52 | 40.95 | 32.48 | 0.7 |
| freecell-3 | 153.71 | 122.00 | 63.47 | 37.94 | 131.30 | 57.14 | 47.74 | 45.07 | 17 |
| depots-7 | 319.48 | 138.30 | 67.24 | 49.58 | 167.24 | 66.89 | 48.32 | 42.68 | 3.6 |
| driverlog-11 | 334.28 | 99.37 | 89.73 | 104.87 | 152.08 | 61.63 | 42.81 | 34.70 | 9.7 |
| gripper-8 | 569.26 | 351.87 | 236.93 | 166.19 | 243.44 | 101.11 | 70.84 | 59.18 | 1.1 |

Table 1: Wall time on STRIPS planning problems.

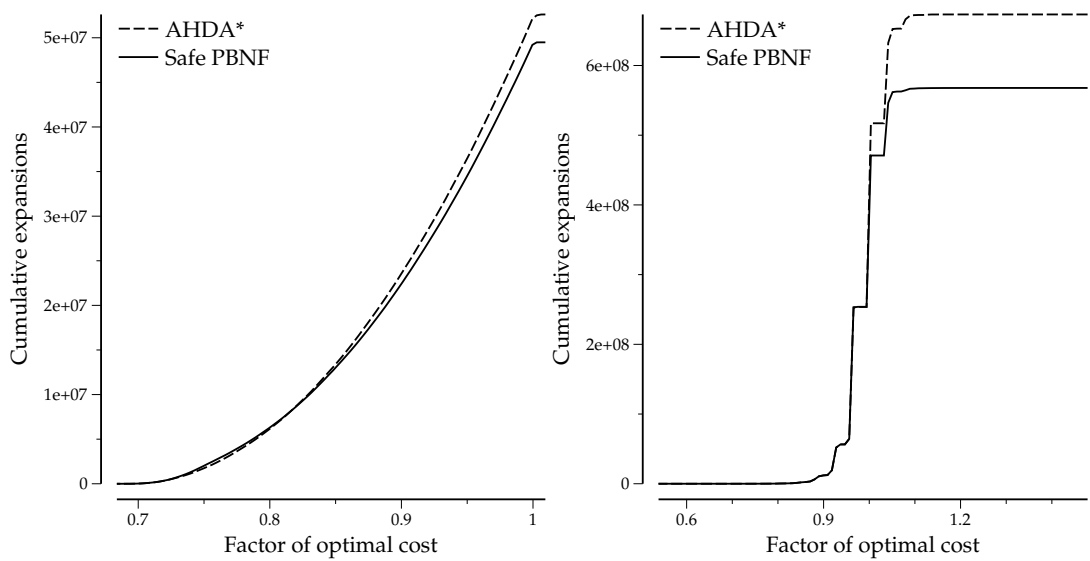

Figure 11: Cumulative normalized $f$ value counts for nodes expanded with eight threads on unit-cost four-way grid pathfinding (left) and the 15-puzzle (right).

three hypotheses. The results of these experiments agree with the first two hypotheses, however, it appears that the third hypothesis does not hold and, in fact, PBNF occasionally spends more time coordinating between threads than AHDA*.

### 4.6.1 NODE QUALITY

Because both PBNF and AHDA* merely approximate a best-first order, they may expand some nodes that have $f$ values greater than the optimal solution cost. When a thread expands a node with an $f$ value greater than the optimal solution cost its effort was a waste because the only nodes that must be expanded when searching for an optimal solution are those with $f$ values less than the optimal cost. In addition to this, both search algorithms may re-expand nodes for which a lower cost path has been found. If this happens work was wasted during the first sub-optimal expansion of the node.

Threads in PBNF are able to choose which $n$block to expand based on the quality of nodes in the free $n$blocks. In AHDA*, however, a thread must expand only those nodes that are assigned to it. We hypothesized that PBNF may expand fewer nodes with $f$ values that are greater than the optimal solution cost because the threads have more control over the quality of the nodes that they choose to expand.

We collected the $f$ value of each node expanded by both PBNF and AHDA*. Figure 11 shows cumulative counts for the $f$ values of nodes expanded by both PBNF and AHDA* on the same set of unit-cost four-way 5000x5000 grid pathfinding instances as were used in Section 4.3 (right) and on the 15-puzzle instances used in Section 4.4 (left). In both plots, the x axis shows the $f$ value of expanded nodes as a factor of the optimal solution cost for the given instance. The y axis shows the cumulative count of nodes expanded up to the given normalized $f$ over the set of instances.

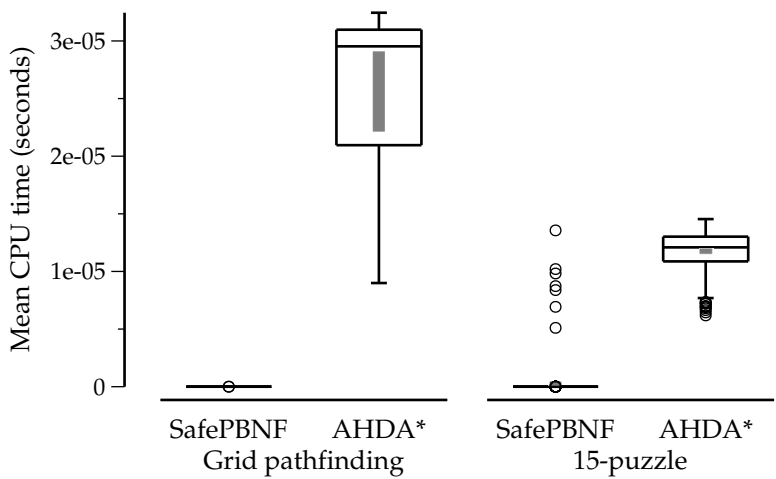

Figure 12: Mean CPU time per open list operation.

By looking at y-location of the right-most tip of each line we can find the total number of nodes expanded by each algorithm summed over all instances.

On the left panel of Figure 11 we can see that both algorithms tended to expand only a very small number of nodes with $f$ values that were greater than the optimal solution cost on the grid pathfinding domain. The AHDA* algorithm expanded more nodes in total on this set of instances. Both PBNF and AHDA* must expand all of the nodes below the optimal solution cost. Because of this, the only way that AHDA* can have a greater number of expansions for nodes below a factor of 1 is if it re-expanded nodes. It appears that AHDA* re-expanded more nodes than PBNF and this seems to account for the fact that AHDA* expanded more nodes in total.

The right half of Figure 11 shows the results on the 15-puzzle. We see that, again, AHDA* expanded more nodes in total than PBNF. In this domain the algorithms expanded approximately the same number of nodes with $f$ values less than the optimal solution cost. We can also see from this plot that AHDA* expanded many more nodes that had $f$ values greater than or equal to the optimal solution cost. In summary, PBNF expanded fewer nodes and better quality nodes than AHDA* in both the grid pathfinding and sliding tiles domains. We speculate that this may happen because in PBNF the threads are allowed to choose which portion of the space they search and they choose it based on low $f$ value. In AHDA* the threads must search the nodes that map to them and these nodes may not be very good.

### 4.6.2 OPEN LIST SIZES

We have found that, since PBNF breaks up the search space into many different $n$blocks, it tends to have data structures with many fewer entries than AHDA*, which breaks up the search space based on the number of threads. Since we are interested general-purpose algorithms that can handle domains with real-valued costs (like eight-way grid pathfinding) both PBNF and AHDA* use binary heaps to implement their open lists. PBNF has one heap per $n$block (that is one per abstract state) whereas AHDA* has one heap per thread. Because the number of $n$blocks is greater than the number of threads AHDA* will have many more nodes than PBNF in each of its heaps. This causes the heap operations in AHDA* to take longer than the heap operations in PBNF.

The cost of operations on large heaps has been shown to greatly impact overall performance of an algorithm (Dai & Hansen, 2007). In order to determine the extent to which large heaps effect the performance of AHDA* we added timers to all of the heap operations for both algorithms. Figure 12 shows the mean CPU time for a single open list operation for unit-cost four-way grid pathfinding domain and for the 15-puzzle. The boxes show the second and third quartiles with a line drawn across at the median. The whiskers show the extremes of the data except that data points residing beyond the first and third quartile by more than 1.5 times the inter-quartile range are signified by a circle. The shaded rectangle shows the 95% confidence interval on the mean. We can see that, in both cases, AHDA* tended to spend more time performing heap operations than PBNF which typically spent nearly no time per heap operation. Heap operations must be performed once for each node that is expanded and may be required on each node generation. Even though these times are in the tens of microseconds the frequency of these operations can be very high during a single search.

Finally, as is described by Hansen and Zhou (2007), the reduction in open list sizes can also explain the good single thread performance that PBNF experiences on STRIPS planning (see Table 1). Hansen and Zhou point out that, although A* is optimally efficient in terms of node expansions, it is not necessarily optimal with respect to wall time. They found that the benefit of managing smaller open lists enabled the Anytime weighted A* algorithm to outperform A* in wall time even though it expanded more nodes when converging to the optimal solution. As we describe in Section 9, this good single thread performance may also be caused by speculative expansions and pruning.

### 4.6.3 COOORDINATION OVERHEAD

Our third hypothesis was that the amount of time that each algorithm spent on "coordination overhead" might differ. Both parallel algorithms must spend some of their time accessing data structures shared among multiple threads. This can cause overhead in two places. The first place where coordination overhead can be seen is in the synchronization of access to shared data structures. PBNF has two modes of locking the $n$block graph. First, if a thread has ownership of an $n$block with open nodes that remain to be expanded then it will use `try_lock` because there is work that could be done if it fails to acquire the lock. Otherwise, if there are no nodes that the thread could expand then it attempt to acquire the lock on the $n$block graph using the normal operation that blocks on failure. AHDA* will use a `try_lock` on its receive queue at each expansion where it has nodes on this queue and on its open list. In our implementation AHDA* will only use the blocking lock operation when a thread has no nodes remaining to expand but has nodes remaining in its send or receive buffers.

The second place where overhead may be incurred is when threads have no nodes to expand. In PBNF this occurs when a thread exhausts its current $n$block and there are no free $n$blocks to acquire. The thread must wait until a new $n$block becomes free. In AHDA* if no open nodes map to a thread then it may have no nodes to expand. In this situation the thread will busy-wait until a node arrives on its receive queue. In either situation, locking or waiting, there is time that is wasted because threads are not actively searching the space.

When evaluating coordination overhead, we combine the amount of time spent waiting on a lock and the amount of time waiting without any nodes to expand. Figure 13 shows the per-thread coordination times for locks, waiting and the sum of the two normalized to the total wall time.

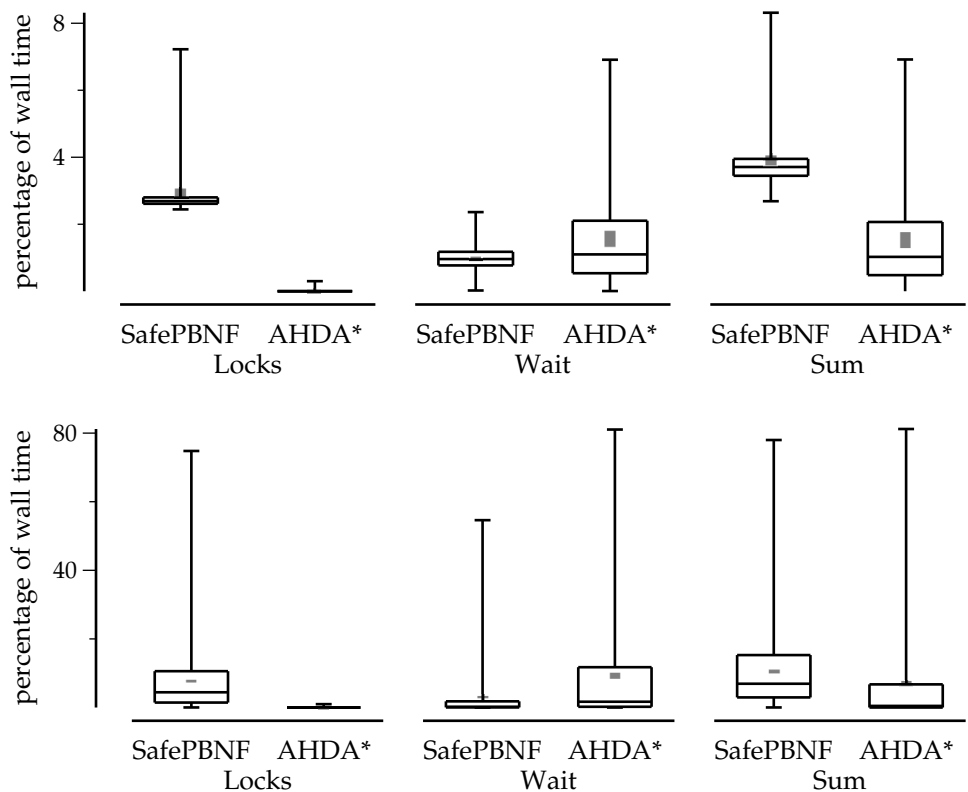

Figure 13: Per-thread ratio of coordination time to wall time on unit-cost four-way pathfinding (top) and the 15-puzzle (bottom).

Unlike the previous set of boxplots, individual data points residing at the extremes are not signified by circles in order to improve readability. The "Locks" column of this plot shows the distribution of times spent by each thread waiting on a lock, the "Wait" column shows the distribution of times that threads spent waiting without any nodes available to expand and the "Sum" column shows the distribution of the sum of the mean lock and wait times.

The left side of Figure 13 shows the results for grid pathfinding. From "Locks" column we see that threads in AHDA* spent almost no time acquiring locks. This is expected because AHDA* uses asynchronous communication. It appears that the amount of time that threads in PBNF spent acquiring locks was significantly greater than that of AHDA*. The "Wait" column of this plot shows that both PBNF and AHDA* appeared to have threads spend nearly the same amount of time waiting without any nodes to expand. Finally, the "Sum" column shows that the threads in PBNF spent more time overall coordinating between threads.

The bottom half of Figure 13 shows the coordination overhead for the 15-puzzle domain. Again, we see that threads in AHDA* spent almost no time acquiring a lock. Individual threads in PBNF, however, tended to spend a larger fraction of their time waiting on locks in the sliding tiles domain than in grid pathfinding. In the "Wait" column of this figure we can see that AHDA* spent more time without any nodes to expand than PBNF. Finally, we see that, over all, PBNF spent more time coordinating between threads than AHDA*.

Overall our experiments have verified that our first two hypotheses that PBNF expanded better quality nodes than AHDA* and that it spent less time performing priority queue operations than AHDA*. We also found that our third hypothesis did not hold and that threads in PBNF tended to have more coordination overhead that AHDA* but this seems to be out-weighed by the other two factors.

### 4.7 Summary

In this section we have shown the results of an empirical evaluation of optimal parallel best-first search algorithms. We have shown that several simple parallel algorithms can actually be slower than a serial A* search even when offered more computing power. Additionally we showed empirical results for a set of algorithms that make good use of parallelism and do outperform serial A*. Overall the Safe PBNF algorithm gave the best and most consistent performance of this latter set of algorithms. Our AHDA* variant of PRA* had the second fastest mean performance in all domains.

We have also shown that using abstraction in a PRA* style search to distribute nodes among the different threads can give a significant boost in speed by reducing the amount of communication. This modification to PRA* appears to be a lot more helpful than simply using asynchronous communication. Using both of these improvements in conjunction (AHDA*), yields a competitive algorithm that has the additional feature of not relying on shared memory.

Finally, we performed a set of experiments in an attempt to explain why Safe PBNF tended to give better search performance than AHDA*. Our experiments looked at three factors: node quality, open list sizes and thread-coordination overhead. We concluded that PBNF is faster because it expands fewer nodes with suboptimal $f$ values and it takes less time to perform priority queue operations.

## 5. Bounded Suboptimal Search

Sometimes it is acceptable or even preferable to search for a solution that is not optimal. Suboptimal solutions can often be found much more quickly and with lower memory requirements than optimal solutions. In this section we show how to create bounded-suboptimal variants of some of the best optimal parallel search algorithms.

Weighted A* (Pohl, 1970), a variant of A* that orders its search on $f'(n) = g(n) + w \cdot h(n)$, with $w > 1$, is probably the most popular suboptimal search. It guarantees that, for an admissible heuristic $h$ and a weight $w$, the solution returned will be $w$-admissible (within a $w$ factor of the optimal solution cost) (Davis, Bramanti-Gregor, & Wang, 1988).

It is possible to modify AHDA*, BFPSDD, and PBNF to use weights to find suboptimal solutions, we call these algorithms wAHDA*, wBFPSDD and wPBNF. Just as in optimal search, parallelism implies that a strict $f'$ search order will not be followed. The proof of weighted A*'s $w$-optimality depends crucially on following a strict $f'$ order, and for our parallel variants we must prove the quality of our solution by either exploring or pruning all nodes. Thus finding effective pruning rules can be important for performance. We will assume throughout that $h$ is admissible.

## 5.1 Pruning Poor Nodes

Let $s$ be the current incumbent solution and $w$ the suboptimality bound. A node $n$ can clearly be pruned if $f(n) \geq g(s)$. But according to the following theorem, we only need to retain $n$ if it is on the optimal path to a solution that is a factor of $w$ better than $s$. This is a much stronger rule.

**Theorem 4** *We can prune a node $n$ if $w \cdot f(n) \geq g(s)$ without sacrificing $w$-admissibility.*

**Proof:** If the incumbent is $w$-admissible, we can safely prune any node, so we consider the case where $g(s) > w \cdot g(opt)$, where *opt* is an optimal goal. Note that without pruning, there always exists a node $p$ in some open list (or being generated) that is on the best path to *opt*. Let $f^*$ be the cost of an optimal solution. By the admissibility of $h$ and the definition of $p$, $w \cdot f(p) \leq w \cdot f^*(p) = w \cdot g(opt)$. If the pruning rule discards $p$, that would imply $g(s) \leq w \cdot f(p)$ and thus $g(s) \leq w \cdot g(opt)$, which contradicts our premise. Therefore, an open node leading to an optimal solution will not be pruned if the incumbent is not $w$-admissible. A search that does not terminate until open is empty will not terminate until the incumbent is $w$-admissible or it is replaced by an optimal solution. □

We make explicit a useful corollary:

**Corollary 2** *We can prune a node $n$ if $f'(n) \geq g(s)$ without sacrificing $w$-admissibility.*

**Proof:** Clearly $w \cdot f(n) \geq f'(n)$, so Theorem 4 applies. □

With this corollary, we can use a pruning shortcut: when the open list is sorted on increasing $f'$ and the node at the front has $f' \geq g(s)$, we can prune the entire open list.

## 5.2 Pruning Duplicate Nodes

When searching with an inconsistent heuristic, as in weighted A*, it is possible for the search to find a better path to an already-expanded state. Likhachev, Gordon, and Thrun (2003) noted that, provided the underlying heuristic function $h$ is consistent, weighted A* will still return a $w$-admissible solution if these duplicate states are pruned during search. This ensures that each state is expanded at most once during the search. Unfortunately, their proof depends on expanding in exactly best-first order, which is violated by several of the parallel search algorithms we consider here. However, we can still prove that some duplicates can be dropped. Consider the expansion of a node $n$ that re-generates a duplicate state $d$ that has already been expanded. We propose the following weak duplicate dropping criterion: the new copy of $d$ can be pruned if the old $g(d) \leq g(n) + w \cdot c^*(n, d)$, where $c^*(n, d)$ is the optimal cost from node $n$ to node $d$.

**Theorem 5** *Even if the weak dropping rule is applied, there will always be a node $p$ from an optimal solution path on* open *such that $g(p) \leq w \cdot g^*(p)$.*

**Proof:** We proceed by induction over iterations of search. The theorem clearly holds after expansion of the initial state. For the induction step, we note that node $p$ is only removed from *open* when it is expanded. If its child $p_i$ that lies along the optimal path is added to *open*, the theorem holds. The only way it won't be added is if there exists a previous duplicate copy $p'_i$ and the pruning rule holds, i.e., $g(p'_i) \leq g(p_{i-1}) + w \cdot c^*(p_{i-1}, p_i)$. By the inductive hypothesis, $g(p_{i-1}) \leq w \cdot g^*(p_{i-1})$, and by definition $g^*(p_{i-1}) + c^*(p_{i-1}, p_i) = g^*(p_i)$, so we have $g(p'_i) \leq w \cdot g^*(p'_i)$. □

Note that the use of this technique prohibits using the global minimum $f$ value as a lower bound on the optimal solution's cost, because $g$ values can now be inflated by up to a factor of $w$. However, if $s$ is the incumbent and we search until the global minimum $f'$ value is $\geq g(s)$, as in a serial weighted A* search, then $w$-admissibility is assured:

**Corollary 3** *If the minimum $f'$ value is $\geq g(s)$, where $s$ is the incumbent, then we have $g(s) \leq w \cdot g^*(\text{opt})$*

***Proof:*** Recall node $p$ from Theorem 5. $g(s) \leq f'(p) = g(p) + w \cdot h(p) \leq w \cdot (g^*(p) + h(p)) \leq w \cdot g^*(opt)$. □

It remains an empirical question whether pruning on this rather weak criterion will lead to better performance in practice. Our results indicate that it does provide an advantage in the grid pathfinding domain. Results are presented in Section 6.1. It should be noted that, while extra pruning can preserve $w$-admissibility, it may result in solutions of lower quality than those resulting from search without pruning.

### 5.3 Optimistic Search

Korf (1993) showed that weighted A* typically returns solutions that are better than the bound, $w$, would suggest. To take advantage of this, Thayer and Ruml (2008) use an optimistic approach to bounded suboptimal search that works in two stages: aggressive search using a weight that is greater than the desired optimality bound to find an incumbent solution and then a cleanup phase to prove that the incumbent is indeed within the bound. The intuition behind this approach is that wA* can find a solution within a very tight bound (much tighter than $w \cdot g(opt)$), then the search can continue looking at nodes in $f$ order until the bound can be proved. Thayer and Ruml show that, indeed, this approach can surpass the speed of wA* for a given optimality bound. We have implemented an optimistic version of PBNF (oPBNF).

One of the requirements of oPBNF is that it must have access to the minimum $f$ value over all nodes in order to prove the bound on the incumbent solution. For the aggressive search stage, the open lists and the heap of free $n$blocks are sorted on $f'$ instead of $f$ so a couple of additions need to be made. First, each $n$block has an additional priority queue containing the open search nodes sorted on $f$. We call this queue $open_f$. The $open_f$ queue is simply maintained by adding and removing nodes as nodes are added and removed from the $f'$ ordered open list of each $n$block. Second, a priority queue, called $\min_f$, of all of the $n$blocks is maintained, sorted on the lowest $f$ value in each $n$block at the time of its last release. $\min_f$ is used to track a lower bound on the minimum $f$ value over all nodes. This is accomplished by lazily updating $\min_f$ only when an $n$block is released by a thread. When a thread releases an $n$block, it sifts the released $n$block and its successors to their new positions in the $\min_f$ queue. These are the only $n$blocks whose minimum $f$ values could have been changed by the releasing thread. Since the global minimum $f$ value over all nodes is strictly increasing (assuming a consistent heuristic) we have the guarantee that the $f$ value at the front of the $\min_f$ queue is strictly increasing and is a lower bound on the global minimum $f$ value at any given time. Using this lower bound, we are able to prove whether or not an incumbent solution is properly bounded.

oPBNF needs to decide when to switch between the aggressive search phase and the cleanup phase of optimistic search. As originally proposed, optimistic search performs aggressive search until the first incumbent is found then it switches between cleanup (when $f'(n) \geq g(s)$, where $n$ is the best node based on $f'$ and $s$ is the incumbent solution) and aggressive search (when $f'(n) < g(s)$) to hedge against the case when the current incumbent is not within the bound. In oPBNF, we were left with a choice: switch between aggressive search and cleanup on a global basis or on a per-$n$block basis. We choose to switch on a per-$n$block basis under the assumption that some threads could be cleaning up areas of the search space with low $f$ values while other threads look for better solutions in areas of the search space with low $f'$ values. In oPBNF, when deciding if one $n$block is better than another (when deciding to switch or to set an $n$block to hot), the choice is no longer based solely on the best $f'$ value of the given $n$block, but instead it is based on the $f'$ value first, then the $f$ value to break ties of if the best $f'$ value is out of the bound of the incumbent. When acquiring a new $n$block, a thread takes either the free $n$block with the best $f'$ value or best $f$ value depending on which $n$block is better (where the notion of better is described in the previous sentence). Finally, when expanding nodes, a thread selects aggressive search or cleanup based on the same criteria as standard optimistic search for the nodes within the acquired $n$block.

## 6. Empirical Evaluation: Bounded Suboptimal Search

We implemented and tested weighted versions of the parallel search algorithms discussed above: wAHDA*, wAPRA*, wBFPSDD, wPBNF and oPBNF. All algorithms prune nodes based on the $w \cdot f$ criterion presented in Theorem 4 and prune entire open lists on $f'$ as in Corollary 2. Search terminates when all nodes have been pruned by the incumbent solution. Our experiments were run on the same three benchmark domains as for optimal search: grid pathfinding, the sliding tile puzzle, and STRIPS planning.

### 6.1 Grid Pathfinding

Results presented in Table 2 show the performance of the parallel search algorithms in terms of speedup over serial weighted A* on grid pathfinding problems. Duplicate states that have already been expanded are dropped in the serial wA* algorithm, as discussed by Likhachev et al. (2003).

The rows of this table show the number of threads and different algorithms whereas the columns are the weights used for various domains. Each entry shows the mean speedup over serial weighted A*. We performed a Wilcoxon signed-rank test to determine which mean values were significantly different; elements that are in **bold** represent values that were not significantly different ($p < 0.05$) from the best mean value in the given column. In general, the parallel algorithms show increased speedup as threads are added for low weights, and decreased speedup as the weight is increased.

In unit-cost four-way movement grids, for weights of 1.1, and 1.2 the wPBNF algorithm was the fastest of all of the algorithms tested reaching over five times the speed of wA* at a weight of 1.1 at and over 4.5x at a weight of 1.2 . At a weight of 1.4 wPBNF, wBFPSDD and wAHDA* did not show a significant difference in performance at 8 threads. wAHDA* had the best speed up of all algorithms at a weight of 1.8. wAPRA* never gave the best performance in this domain.

In eight-way movement grids wPBNF gave the best performance for a weight of 1.1 and 1.4, although in the latter case this best performance was a decrease over the speed of wA* and it was achieved at 1 thread. wAHDA* was the fastest when the weight was 1.2, however, this did not scale as expected when the number of threads was increased. Finally wAPRA* gave the least performance decrease over weighted A* at a weight of 1.8 with 1 thread. In this case, all algorithms were slower than serial weighted A* but wAPRA* gave the closest performance to the serial search. wBFPSDD never gave the best performance in this domain.

In the life-cost domain wPBNF outperformed all other algorithms for weights 1.1, 1.2 and 1.4. At weight 1.8, wPBNF's performance quickly dropped, however and wAHDA* had the best results with more than a 4x speedup over wA*, although the performance appears to have been very inconsistent as it is not significantly different from much lower speedup values for the same weight. wAPRA* never gave the best performance in this domain.

|  |  | weight | | | | | | | | | | | |
|  |  | Unit Four-way Grids | | | | Unit Eight-way Grids | | | | Life Four-way Grids | | | |
|  |  | 1.1 | 1.2 | 1.4 | 1.8 | 1.1 | 1.2 | 1.4 | 1.8 | 1.1 | 1.2 | 1.4 | 1.8 |
|---|---|---|---|---|---|---|---|---|---|---|---|---|---|
| wPBNF | 1 | 0.98 | 0.91 | 0.51 | 0.73 | 0.93 | 1.37 | **0.73** | 0.74 | 0.65 | 0.66 | 0.84 | 0.67 |
|  | 2 | 1.74 | 1.65 | 1.07 | 0.87 | 1.65 | 1.82 | 0.57 | 0.66 | 1.15 | 1.17 | 1.59 | 0.39 |
|  | 3 | 2.47 | 2.33 | 1.62 | 0.89 | 2.36 | 1.77 | 0.55 | 0.61 | 1.65 | 1.67 | 2.32 | 0.39 |
|  | 4 | 3.12 | 2.92 | 2.13 | 0.90 | 2.97 | 1.72 | 0.53 | 0.58 | 2.08 | 2.10 | 2.96 | 0.49 |
|  | 5 | 3.76 | 3.52 | 2.48 | 0.91 | 3.55 | 1.67 | 0.52 | 0.56 | 2.53 | 2.55 | 3.63 | 1.49 |
|  | 6 | 4.30 | 3.99 | 2.80 | 0.89 | 4.04 | 1.61 | 0.50 | 0.54 | 2.94 | 2.95 | 4.20 | 1.64 |
|  | 7 | 4.78 | 4.40 | **3.01** | 0.88 | 4.40 | 1.55 | 0.49 | 0.51 | 3.31 | 3.33 | 4.63 | 2.12 |
|  | 8 | **5.09** | **4.66** | **3.11** | 0.87 | **4.70** | 1.49 | 0.45 | 0.46 | **3.61** | **3.64** | **5.11** | 1.06 |
| wBFPSDD | 1 | 0.82 | 0.84 | 0.96 | 0.94 | 0.87 | 0.79 | 0.43 | 0.33 | 0.52 | 0.53 | 0.58 | 0.60 |
|  | 2 | 1.26 | 1.26 | 1.45 | 0.91 | 1.37 | 1.10 | 0.43 | 0.35 | 0.83 | 0.83 | 0.92 | 0.76 |
|  | 3 | 1.65 | 1.65 | 1.90 | 0.84 | 1.80 | 1.22 | 0.41 | 0.33 | 1.10 | 1.09 | 1.26 | 0.84 |
|  | 4 | 1.93 | 1.92 | 2.09 | 0.79 | 2.13 | 1.25 | 0.42 | 0.33 | 1.29 | 1.29 | 1.48 | 0.89 |
|  | 5 | 2.24 | 2.24 | 2.36 | 0.75 | 2.47 | 1.31 | 0.39 | 0.32 | 1.53 | 1.51 | 1.61 | **0.93** |
|  | 6 | 2.51 | 2.51 | **2.58** | 0.71 | 2.74 | 1.21 | 0.36 | 0.30 | 1.73 | 1.72 | 1.78 | 0.93 |
|  | 7 | 2.73 | 2.69 | 2.63 | 0.67 | 2.94 | 1.26 | 0.34 | 0.29 | 1.91 | 1.89 | 1.94 | 0.91 |
|  | 8 | 2.91 | 2.84 | **2.68** | 0.63 | 3.10 | 1.23 | 0.32 | 0.26 | 2.06 | 2.03 | 2.10 | 0.85 |
| wAHDA* | 1 | 0.87 | 0.79 | 0.32 | 0.56 | 0.79 | 1.10 | 0.66 | 0.76 | 0.56 | 0.55 | 0.71 | 0.22 |
|  | 2 | 1.35 | 1.17 | 0.63 | 0.84 | 1.04 | 1.99 | 0.62 | 0.61 | 0.88 | 0.86 | 1.29 | 0.32 |
|  | 3 | 1.90 | 1.69 | 1.30 | **1.30** | 2.08 | **2.93** | 0.64 | 0.62 | 1.09 | 1.39 | 1.86 | 0.56 |
|  | 4 | 2.04 | 2.10 | 1.57 | **1.30** | 2.48 | **2.84** | 0.56 | 0.54 | 1.60 | 1.64 | 2.24 | 0.56 |
|  | 5 | 1.77 | 2.08 | 1.79 | 0.97 | 2.49 | 2.52 | 0.42 | 0.41 | 1.88 | 1.92 | 2.58 | 0.41 |
|  | 6 | 3.23 | 3.03 | 2.18 | **1.33** | 3.73 | **2.83** | 0.49 | 0.45 | 2.15 | 2.17 | 3.02 | **1.50** |
|  | 7 | 3.91 | 3.78 | 2.56 | **1.30** | 4.45 | **2.89** | 0.45 | 0.41 | 2.39 | 2.41 | 3.50 | **1.07** |
|  | 8 | 3.79 | 3.64 | **3.02** | 1.13 | 4.39 | 2.58 | 0.37 | 0.38 | 2.38 | 2.42 | 3.55 | **4.16** |
| wAPRA* | 1 | 0.88 | 0.81 | 0.32 | 0.56 | 0.80 | 1.11 | 0.67 | **0.77** | 0.56 | 0.56 | 0.72 | 0.23 |
|  | 2 | 0.51 | 0.44 | 0.22 | 0.36 | 0.35 | 0.69 | 0.31 | 0.28 | 0.35 | 0.34 | 0.46 | 0.12 |
|  | 3 | 0.36 | 0.32 | 0.20 | 0.26 | 0.41 | 0.65 | 0.23 | 0.22 | 0.23 | 0.26 | 0.32 | 0.10 |
|  | 4 | 0.50 | 0.44 | 0.30 | 0.41 | 0.43 | 0.73 | 0.22 | 0.19 | 0.42 | 0.43 | 0.55 | 0.16 |
|  | 5 | 0.55 | 0.56 | 0.39 | 0.48 | 0.49 | 0.87 | 0.23 | 0.19 | 0.54 | 0.56 | 0.67 | 0.20 |
|  | 6 | 0.52 | 0.49 | 0.31 | 0.30 | 0.50 | 0.65 | 0.16 | 0.14 | 0.39 | 0.39 | 0.49 | 0.13 |
|  | 7 | 0.73 | 0.67 | 0.40 | 0.36 | 0.62 | 0.73 | 0.17 | 0.14 | 0.49 | 0.49 | 0.65 | 0.18 |
|  | 8 | 1.09 | 1.07 | 0.82 | 0.77 | 0.89 | 1.38 | 0.28 | 0.22 | 1.00 | 0.98 | 1.22 | 0.42 |

*(Row group label on the far left spanning all rows: threads)*

Table 2: Grid Pathfinding: Average speedup over serial weighted A* for various numbers of threads.

| threads | wPBNF | | | | wBFPSDD | | | |
|---|---|---|---|---|---|---|---|---|
| | 1.4 | 1.7 | 2.0 | 3.0 | 1.4 | 1.7 | 2.0 | 3.0 |
| 1 | 0.68 | 0.44 | 0.38 | 0.69 | 0.65 | 0.61 | 0.44 | 0.35 |
| 2 | 1.35 | 0.81 | 1.00 | 0.63 | 0.87 | 0.74 | 0.49 | 0.43 |
| 3 | 1.48 | 0.97 | 0.85 | 0.56 | 1.05 | 0.72 | 0.63 | 0.46 |
| 4 | 1.70 | 1.20 | 0.93 | 0.60 | 1.09 | 1.00 | 0.57 | 0.45 |
| 5 | 2.04 | 1.38 | 0.97 | 0.74 | 1.27 | 0.97 | 0.65 | 0.40 |
| 6 | 2.16 | 1.30 | **1.19** | 0.67 | 1.33 | 1.17 | 0.61 | 0.39 |
| 7 | **2.55** | **1.46** | 1.04 | 0.62 | 1.49 | 1.10 | 0.59 | 0.34 |
| 8 | **2.71** | **1.71** | 1.10 | 0.60 | 1.53 | 1.08 | 0.62 | 0.33 |

| threads | wAHDA* | | | | wAPRA* | | | |
|---|---|---|---|---|---|---|---|---|
| | 1.4 | 1.7 | 2.0 | 3.0 | 1.4 | 1.7 | 2.0 | 3.0 |
| 1 | 0.61 | 0.60 | 0.59 | 0.54 | 0.61 | 0.59 | 0.59 | 0.54 |
| 2 | 1.18 | 1.11 | 1.32 | **0.78** | 1.18 | 1.08 | 1.36 | 0.78 |
| 3 | 1.53 | **1.30** | 1.40 | 0.73 | 1.45 | 1.25 | 1.32 | 0.78 |
| 4 | 1.91 | **1.57** | 1.55 | 0.74 | 1.77 | **1.50** | 1.36 | 0.62 |
| 5 | **2.33** | 1.70 | 1.27 | 0.66 | 2.32 | 1.62 | 1.26 | 0.64 |
| 6 | **2.28** | **1.72** | 1.24 | 0.52 | 2.18 | 1.54 | **1.83** | 0.47 |
| 7 | 2.71 | 1.50 | 1.03 | 0.44 | 2.63 | 1.40 | 1.09 | 0.43 |
| 8 | **2.70** | **1.51** | 1.24 | 0.44 | **2.34** | **1.61** | 1.22 | 0.41 |

Table 3: 15-puzzle: Average speedup over serial weighted A* for various numbers of threads.

| | threads | Unit Four-way Grids | | | | Unit Eight-way Grids | | | | 250 easy 15-puzzles | | | |
|---|---|---|---|---|---|---|---|---|---|---|---|---|---|
| | | 1.1 | 1.2 | 1.4 | 1.8 | 1.1 | 1.2 | 1.4 | 1.8 | 1.4 | 1.7 | 2.0 | 3.0 |
| | 1 | 0.54 | 0.99 | 0.74 | **0.47** | 0.74 | **0.76** | **0.09** | **0.05** | 0.56 | 0.58 | 0.77 | 0.60 |
| | 2 | 0.99 | 2.00 | 1.05 | 0.45 | 1.26 | 0.71 | 0.09 | 0.05 | 0.85 | **1.07** | **0.83** | **0.72** |
| oPBNF | 3 | 1.40 | 2.89 | 1.19 | 0.45 | 1.64 | 0.70 | 0.09 | 0.05 | 1.06 | **0.94** | **0.79** | **0.80** |
| | 4 | 1.76 | 3.62 | 1.26 | 0.44 | 1.90 | 0.69 | 0.09 | 0.05 | 1.01 | 0.82 | **0.93** | **0.69** |
| | 5 | 2.11 | 4.29 | 1.33 | 0.43 | 2.09 | 0.68 | 0.08 | 0.05 | **1.20** | **1.21** | **0.97** | **0.74** |
| | 6 | 2.43 | 4.84 | **1.35** | 0.44 | 2.21 | 0.68 | 0.08 | 0.05 | **1.32** | **0.83** | **0.99** | 0.67 |
| | 7 | 2.70 | 5.44 | **1.37** | 0.43 | 2.29 | 0.67 | 0.08 | 0.04 | **1.14** | **0.93** | **0.88** | 0.71 |
| | 8 | **2.97** | **6.01** | **1.39** | 0.42 | **2.30** | 0.67 | 0.08 | 0.04 | **1.33** | 0.87 | 0.81 | 0.64 |

Table 4: Average speedup over serial optimistic search for various numbers of threads.

Overall, we see that wPBNF often had the best speedup results at eight threads and for weights less than 1.8. wAHDA*, however, gave the best performance at a weight of 1.8 across all grid pathfinding domains. wBFPSDD often gave speedup over serial weighted A*, however it was not quite as competitive as wPBNF or wAHDA*. wAPRA* was only very rarely able to outperform the serial search.

Table 4 shows the results for the optimistic variant of the PBNF algorithm (oPBNF). Each cell in this table shows the mean speedup of oPBNF over serial optimistic search. Once again, the **bold** cells entries that are not significantly different from the best value in the column. For unit-cost four-way pathfinding problems oPBNF gave a performance increase over optimistic search for two or more threads and for all weights less than 1.8. At a weight of 1.2, oPBNF tended to give the best speedup, this may be because optimistic search performed poorly at this particular weight. In unit-cost eight-way pathfinding, we see that oPBNF performs comparably to the unit-cost domain for a weight of 1.1, however, at all higher weights the algorithm is slower than serial optimistic search.

## 6.2 Sliding Tile Puzzles

For the sliding tiles domain, we used the standard Korf 100 15-puzzles (Korf, 1985). Results are presented in Table 3. wPBNF, wAHDA* and wAPRA* tended to give comparable performance in the sliding tile puzzle domain each having values that are not significantly different for weights of 1.4 and 1.7. At a weight of 3.0, wAHDA* gave the least performance decrease over weighted A* at 2 threads.

The right-most column of Table 4 shows the results for optimistic PBNF on 250 15-puzzle instances that were solvable by A* in fewer than 3 million expansions. oPBNF gave its best performance at a weight of 1.4. For weights greater than 1.4 oPBNF was unable to outperform its serial counterpart. For greater weights oPBNF tended to perform better with smaller numbers of threads.

One trend that can be seen in both the sliding tiles domain and the grid pathfinding domain is that the speedup of the parallel algorithms over serial suboptimal search decreases as the weight is increased. We suspect that the decrease in relative performance is due to the problems becoming sufficiently easy (in terms of node expansions) that the overhead for parallelism becomes harmful to overall search. In problems that require many node expansions the cost of parallelism (additional expansions, spawning threads, synchronization – albeit small, waiting for threads to complete, etc.) is amortized by the search effort. In problems that require only a small number of expansions, however, this overhead accounts for more of the total search time and a serial algorithm could potentially be faster.

To confirm our understanding of the effect of problem size on speedup, Figure 14 shows a comparison of wPBNF to weighted A* on all of the 100 Korf 15-puzzle instances using eight threads. Each point represents a run on one instance at a particular weight, the y-axis represents wPBNF speedup relative to serial wA*, and the x-axis represents the number of nodes expanded by wA*. Different glyphs represents different weight values used for both wPBNF and wA*. The figure shows that, while wPBNF did not outperform wA* on easier problems, the benefits of wPBNF over wA* increased as problem difficulty increased. The speed gain for the instances that were run at a weight of 1.4 (the lowest weight tested) leveled off just under 10 times faster than wA*. This is because the machine has eight cores. There are a few instances that seem to have speedup greater than 10x. These can be explained by the speculative expansions that wPBNF performs which may find a bounded solution faster than weighted A* due to the pruning of more nodes with $f'$ values equal to that of the resulting solution. The poor behavior of wPBNF for easy problems is most likely due to the overhead described above. This effect of problem difficulty means that wPBNF outperformed wA* more often at low weights, where the problems required more expansions, and less often at higher weights, where the problems were completed more quickly.

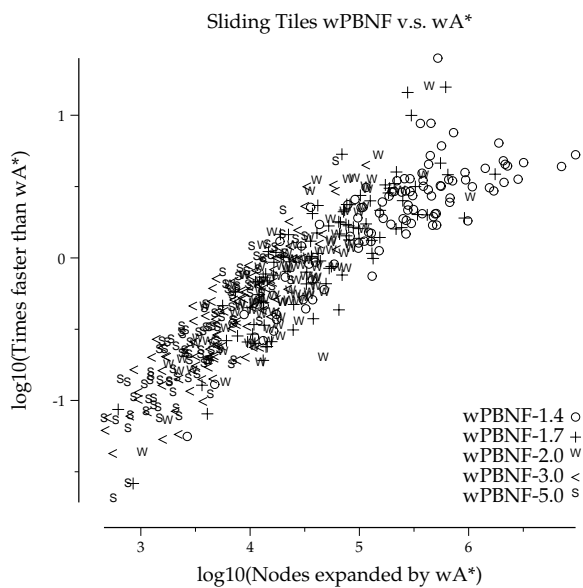

Figure 14: wPBNF speedup over wA* as a function of problem difficulty.

### 6.3 STRIPS Planning

Table 5 shows the performance of the parallel search algorithms on STRIPS planning problems, again in terms of speedup versus serial weighted A*. In this table columns represent various weights and the rows represent different planning problems with two and seven threads. **Bold** values represent table entries that are within 10% of the the best performance for the given domain. All algorithms had better speedup at seven threads than at two. wPBNF gave the best speedup for the most number of domains followed by wAHDA* which was the fastest for three of the domains at seven threads. At two threads there were a couple of domains (satellite-6 and freecell-3) where wBFPSDD gave the most speedup, however it never did at seven threads. wAPRA* was always slower than the three remaining algorithms. On one problem, freecell-3, serial weighted A* performs much worse as the weight increases. Interestingly, wPBNF and wBFPSDD do not show this pathology, and thus record speedups of up to 1,700 times.

### 6.4 Summary

In this section, we have seen that bounded suboptimal variants of the parallel searches can give better performance than their serial progenitors. We have also shown that, on the sliding tile puzzle, parallel search gives more of an advantage over serial search as problem difficulty increases and we suspect that this result holds for other domains too. We suspect that this is because the overhead of using parallelism is not amortized by search time for very easy problems.

| | | wAPRA* | | | | wAHDA* | | | |
|---|---|---|---|---|---|---|---|---|---|
| | | 1.5 | 2 | 3 | 5 | 1.5 | 2 | 3 | 5 |
| **2 threads** | logistics-8 | 0.99 | 1.02 | 0.59 | 1.37 | 1.25 | 1.11 | 0.80 | 1.51 |
| | blocks-16 | 1.29 | 0.88 | 4.12 | 0.30 | 1.52 | 1.09 | **4.86** | 0.38 |
| | gripper-7 | 0.76 | 0.76 | 0.77 | 0.77 | 1.36 | 1.35 | 1.33 | 1.30 |
| | satellite-6 | 0.68 | 0.93 | 0.70 | 0.75 | 1.15 | 1.09 | 1.28 | 1.44 |
| | elevator-12 | 0.65 | 0.72 | 0.71 | 0.77 | 1.16 | 1.20 | 1.27 | 1.22 |
| | freecell-3 | 1.03 | 1.00 | 1.78 | 1.61 | 1.49 | 1.20 | 7.56 | 1.40 |
| | depots-13 | 0.73 | 1.25 | 0.97 | 1.08 | 0.92 | 1.29 | 0.96 | 1.09 |
| | driverlog-11 | 0.91 | 0.79 | 0.94 | 0.93 | **1.30** | 0.97 | 0.96 | 0.93 |
| | gripper-8 | 0.63 | 0.61 | 0.62 | 0.62 | 1.14 | 1.16 | 1.15 | 1.16 |
| **7 threads** | logistics-8 | 3.19 | 3.10 | 3.26 | 2.58 | 4.59 | 4.60 | 3.61 | 2.58 |
| | blocks-16 | 3.04 | 1.37 | 1.08 | 0.37 | **3.60** | 1.62 | 0.56 | 0.32 |
| | gripper-7 | 1.71 | 1.74 | 1.73 | 1.82 | 3.71 | 3.66 | 3.74 | 3.83 |
| | satellite-6 | 1.11 | 1.01 | 1.29 | 1.44 | 3.22 | 3.57 | 3.05 | 3.60 |
| | elevator-12 | 0.94 | 0.97 | 1.04 | 1.02 | 2.77 | 2.88 | 2.98 | 3.03 |
| | freecell-3 | 3.09 | 7.99 | 2.67 | 2.93 | 4.77 | 2.71 | 48.66 | 4.77 |
| | depots-13 | 2.38 | 5.36 | 1.13 | 1.17 | 2.98 | **6.09** | 1.22 | 1.17 |
| | driverlog-11 | 1.90 | 1.25 | 0.93 | 0.92 | **3.52** | 1.48 | 0.95 | 0.92 |
| | gripper-8 | 1.70 | 1.68 | 1.68 | 1.74 | 3.71 | 3.63 | 3.67 | 4.00 |

| | | wPBNF | | | | wBFPSDD | | | |
|---|---|---|---|---|---|---|---|---|---|
| | | 1.5 | 2 | 3 | 5 | 1.5 | 2 | 3 | 5 |
| **2 threads** | logistics-8 | 2.68 | 2.27 | **4.06** | 1.00 | 1.86 | 2.12 | 1.14 | 0.15 |
| | blocks-16 | 0.93 | 0.54 | 0.48 | 1.32 | 0.34 | 0.19 | 0.16 | 0.32 |
| | gripper-7 | **2.01** | **1.99** | **1.99** | **2.02** | **1.91** | **1.89** | **1.86** | **1.84** |
| | satellite-6 | 2.02 | 1.53 | 5.90 | 3.04 | 1.71 | 2.22 | **7.50** | 2.80 |
| | elevator-12 | **2.02** | **2.08** | **2.21** | **2.15** | 1.76 | 1.76 | 1.81 | **2.18** |
| | freecell-3 | 2.06 | 0.84 | 8.11 | 10.69 | 1.42 | 0.54 | 16.88 | **55.75** |
| | depots-13 | 2.70 | **4.49** | 0.82 | 0.81 | 1.48 | 1.58 | 0.18 | 0.14 |
| | driverlog-11 | 0.85 | 0.19 | 0.69 | 0.62 | 0.85 | 0.11 | 0.19 | 0.21 |
| | gripper-8 | **2.06** | **2.04** | **2.08** | **2.07** | **2.00** | **1.96** | **1.97** | **1.98** |
| **7 threads** | logistics-8 | **7.10** | **6.88** | 1.91 | 0.46 | 3.17 | 3.59 | 0.62 | 0.10 |
| | blocks-16 | 2.87 | 0.70 | 0.37 | 1.26 | 0.49 | 0.22 | 0.11 | 0.32 |
| | gripper-7 | **5.67** | 5.09 | 5.07 | **5.18** | 4.33 | 4.28 | 4.14 | 4.05 |
| | satellite-6 | 4.42 | 2.85 | 2.68 | **5.89** | 3.13 | 2.31 | 3.01 | 1.05 |
| | elevator-12 | 6.32 | 6.31 | **6.60** | **7.10** | 3.68 | 3.78 | 4.04 | 3.95 |
| | freecell-3 | 7.01 | 2.31 | 131.12 | **1,721.33** | 2.12 | 0.70 | 44.49 | 137.19 |
| | depots-13 | 3.12 | 1.80 | 0.87 | 0.88 | 1.88 | 1.87 | 0.15 | 0.12 |
| | driverlog-11 | 1.72 | 0.43 | 0.67 | 0.42 | 1.26 | 0.21 | 0.30 | 0.23 |
| | gripper-8 | **5.85** | 5.31 | **5.40** | **5.44** | 4.62 | 4.55 | 4.55 | 4.51 |

Table 5: Speed-up over serial weighted A* on STRIPS planning problems for various weights.

## 7. Anytime Search

A popular alternative to bounded suboptimal search is anytime search, in which a highly suboptimal solution is returned quickly and then improved solutions are returned over time until the algorithm is terminated (or the incumbent solution is proved to be optimal). The two most popular anytime heuristic search algorithms are Anytime weighted A* (AwA*) (Hansen & Zhou, 2007) and anytime repairing A* (ARA*) (Likhachev, Gordon, & Thrun, 2003). In AwA* a weighted A* search is allowed to continue after finding its first solution, pruning when the unweighted $f(n) \geq g(s)$ where $s$ is an incumbent solution and $n$ is a node being considered for expansion. ARA* uses a weighted search where the weight is lowered when a solution meeting the current suboptimality bound has been found and a special *INCONS* list is kept that allows the search to expand a node at most once during the search at each weight.

In this section we present anytime versions of the best performing parallel searches from our previous sections. We used the PBNF framework to implement Anytime weighted PBNF (Aw-PBNF) and Anytime Repairing PBNF (ARPBNF). We use the PRA* framework to create anytime weighted AHDA* (AwAHDA*). We also show the performance of a very simple algorithm that runs parallel weighted A* searches with differing weights. In the planning domain, we have implemented anytime weighted BFPSDD (AwBFPSDD) for comparison as well.

Because our parallel searches inherently continue searching after their first solutions are found, they serve very naturally as anytime algorithms in the style of Anytime weighted A*. The main difference between the standard, optimal versions of these algorithms and their anytime variants is that the anytime versions will sort all open lists and the heap of free $n$blocks on $f'(n) = g(n) + w \cdot h(n)$. In fact, in both cases the optimal search is a degenerate case of the anytime search where $w = 1$. This approach (simply using $w > 1$) is used to implement all algorithms except for ARPBNF and multi-weighted A*.

Next, we will discuss the details of the ARPBNF algorithm. Following that, we introduce a new parallel anytime algorithm called multi-weighted A*. Finally, we show the results of a set of comparisons that we performed on the anytime algorithms discussed in these sections.

### 7.1 Anytime Repairing PBNF

ARPBNF is a parallel anytime search algorithm based on ARA* (Likhachev et al., 2003). In ARPBNF, open lists and the heap of $n$blocks are sorted on $f'$ as in AwPBNF, but instead of merely continuing the search until the incumbent is proved optimal, ARPBNF uses a weight schedule. Each time an incumbent is found, the weight on the heuristic value is lowered by a specified amount, all open lists are resorted and the search continues. On the final iteration, the weight will be $1.0$ and the optimal solution will be found.

The following procedure is used to resort the $n$blocks in parallel between incumbent solutions:

1. The thread calling for a resort (the one that found a goal) becomes the leader by taking the lock on the $n$block graph and setting the *resort flag*. (If the flag has already been set, then another thread is already the leader and the current thread becomes a worker). After the flag is set the leader thread releases the lock on the $n$block graph and waits for all $n$blocks to have $\sigma$ values of zero (no $n$blocks are acquired).

2. Threads check the *resort flag* each expansion, if it is set then threads release their $n$blocks and become worker threads and wait for the leader to set the *start flag*.

3. Once all $n$blocks have $\sigma = 0$, the leader re-takes the lock on the $n$block graph and ensures that all $\sigma$ values are still zero (if not, then it releases the lock and retries). The leader sets the global weight value to the next weight on the weight schedule and populates a lock-free queue with all $n$blocks. Once the queue has been populated, the leader sets the *start flag*.

4. All threads greedily dequeue $n$blocks and resort them until the queue is empty.

5. When all $n$blocks have been resorted, the leader thread clears the *resort flag* and the *start flag* and releases the lock on the $n$block graph. All threads will now acquire new $n$blocks and the search will continue.

We modeled this procedure in TLA$^+$ and showed it to be live-lock and dead-lock free for up to 4 threads and 5 nblocks by the use of the TLC model checker (Yu et al., 1999). This model is very simple so we do not include it in an appendix.

## 7.2 Multi-weighted A*

In this section we introduce a new and simple parallel anytime algorithm called multi-weighted A*. The PBNF and PRA* frameworks for parallelizing anytime algorithms can be thought of as one end on a spectrum of parallel anytime algorithms. In PBNF and PRA* all threads are working on finding a single solution of a given quality; on the opposite end of the spectrum each thread would be working to find its own solution. To compare to an algorithm at that end of the spectrum we implemented an algorithm we call multi-weighted A* that allocates its available threads to their own weighted A* searches. The thread that finishes first will generally be the thread that was searching at the greatest weight and therefore the solution will be of the worst quality. The next thread to finish will have the next greatest weight, and so on. The final thread to complete will generally be searching at a weight of 1.0, performing a standard A* search, and will return the optimal solution.

The algorithm is given a schedule of weighs in decreasing order. The largest weights in the schedule are distributed among the available threads. The threads begin searching using wA* with their given weight values. When a thread finds a new solution that is better than the current one, it updates the incumbent that is shared between all threads to allow for pruning. When a thread finds a better incumbent solution, it will be $w$-admissible with respect to the weight the thread was searching with. If a thread finishes (either finding a solution or pruning its entire open list), it takes the highest unclaimed weight from the schedule and starts a fresh search using that weight. If there are no weights left in the schedule, the thread terminates. When all threads have terminated, the search is complete. If the final weight in the schedule is 1.0, then the last solution found will be optimal.

One of the benefits of multi-weighted A* is that it is a very simple algorithm to implement. However, as we will see below, it doesn't benefit much from added parallelism. A reason for this may be because, when the weight schedule is exhausted (a thread is searching with the lowest weight, 1.0) threads that complete their searches will sit idle until the entire search terminates. Since the final weight will take the longest, this may be a majority of the search time. A more dynamic schedule could be used to keep threads busy until the optimal solution is found. One could also attempt to use more threads at once by using some multi-threaded search at each weight, such as wPBNF or wAHDA*. We leave these extensions for future work.

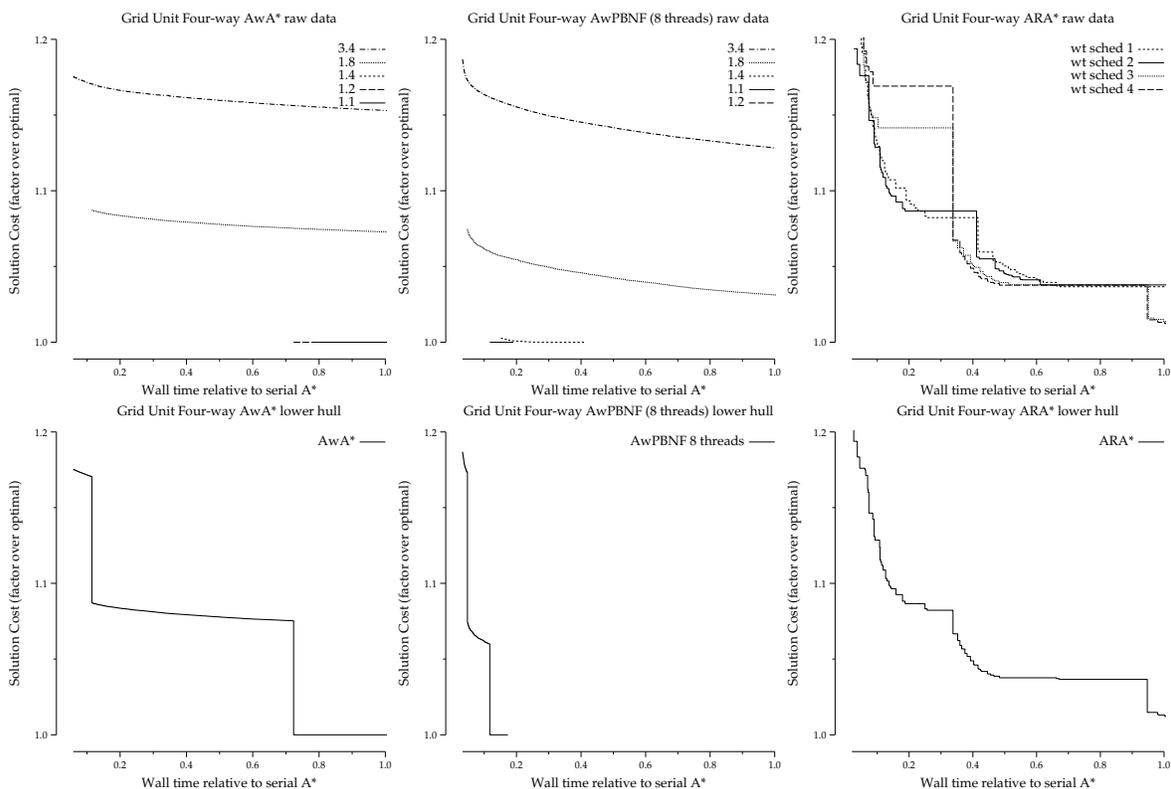

Figure 15: Raw data profiles (top) and lower hull profiles (bottom) for AwA* (left), AwPBNF (center), and ARA* (right). Grid unit-cost four-way pathfinding.

## 8. Empirical Evaluation: Anytime Search

The implementation and empirical setup was similar to that used for suboptimal search. For ARA*, ARPBNF and Multi-wA* we considered four different weight schedules: {7.4, 4.2, 2.6, 1.9, 1.5, 1.3, 1.1, 1}, {4.2, 2.6, 1.9, 1.5, 1.3, 1.1, 1.05, 1}, {3, 2.8, ..., 1.2, 1}, {5, 4.8, ..., 1.2, 1}. For AwA* and the other anytime parallel algorithms we consider weights of: 1.1, 1.2, 1.4, 1.8 and 3.4 for grid pathfinding and 1.4, 1.7, 2.0, 3.0 and 5.0 for the sliding tiles domain. To fully evaluate anytime algorithms, it is necessary to consider their performance profile, i.e., the expected solution quality as a function of time. While this can be easily plotted, it ignores the fact that the anytime algorithms considered in this paper all have a free parameter, namely the weight or schedule of weights used to accelerate the search. In order to compare algorithms, we make the assumption that, in any particular application, the user will attempt to find the parameter setting giving good performance for the timescale they are interested in. Under this assumption, we can plot the performance of each anytime algorithm by computing, at each time point, the best performance that was achieved by any of the parameter settings tried for that algorithm – that is minimum solution cost over all parameter settings for a given algorithm up to the given time point. We refer to this concept as the 'lower hull' of the profiles, because it takes the minimum over the profiles for each parameter setting.

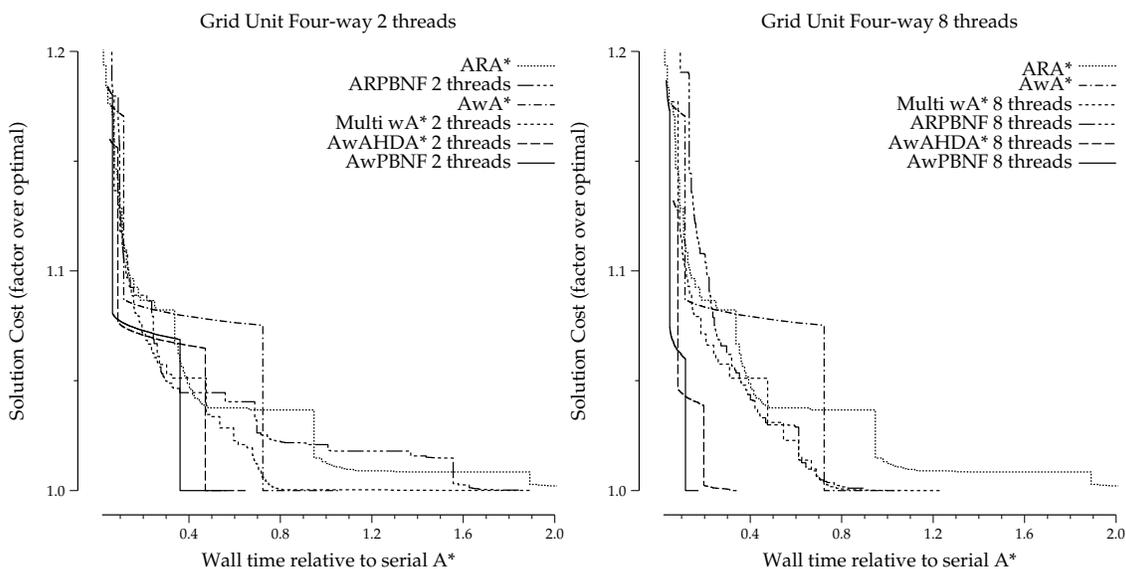

Figure 16: Grid unit-cost four-way pathfinding lower hull anytime profiles.

The top row of Figure 15 shows an example of the raw data for three algorithms on our 5000x5000 unit-cost four-way grid pathfinding problems. The y-axis of these plots is the solution quality as a factor of optimal and the x-axis is the wall clock time relative to the amount of time A* took to find an optimal solution. The bottom row of this figure shows the lower hull for the respective data displayed above. By comparing the two images on the left that display the data for the AwA* algorithm, one can see that the three big "steps" in the lower hull plot is where a different weight is used in the hull because it has found a better solution for the same time bound. The center panel in Figure 15 shows that the AwPBNF algorithm gives a similar performance to AwA*, however it is often faster. This is not surprising since AwPBNF is based on the AwA* approach and it is running at eight threads instead of one. The final panel in Figure 15 shows ARA*, which uses weight schedules instead of a single weight.

Figures 16-17 present the lower hulls of both serial and parallel algorithms on grid pathfinding and the sliding tile puzzle. In each panel, the y-axis represents solution cost as a factor of the optimal cost. In Figure 16 the x-axis represents wall time relative to the amount of time that serial A* took to find an optimal solution. This allows for a comparison between the anytime algorithms and standard serial A*. Since A* is not able to solve all of Korf's 100 15-puzzle instances on this machine, the x-axis in Figure 17 is the absolute wall time in seconds. Both serial and parallel algorithms are plotted. The profiles start when the algorithm first returns a solution and ends when the algorithm has proved optimality or after a 180 second cutoff (since Multi-wA* can consume memory more quickly than the other algorithms, we gave it a 120 second cutoff on the sliding tile puzzle to prevent thrashing).

## 8.1 Four-Way Unit Cost Grids

Figure 16 shows the anytime performance for unit cost four-way movement grid pathfinding problems. AwAHDA* and AwPBNF found the best solutions quicker than the other algorithms. Both of these algorithms improved in the amount of time taken to find better solutions as more threads were added. AwPBNF converged more quickly as more threads were added. Even at two threads AwPBNF was the first algorithm to converge on the optimal solution in 60% of the time of serial A*. The next two algorithms are Multi-wA* and anytime repairing PBNF (ARPBNF). Multi-wA* converged more quickly as threads were added, but its performance on finding intermediate solutions did not change too much for different numbers of threads. ARPBNF, on the other hand, took longer to find good solutions for low thread counts, but as threads were added it started to perform better, eventually matching Multi wA* at eight threads. Both of these algorithms improved the solution quality more steadily than AwPBNF and AwAHDA* which had large jumps in their lower hulls. Each of these jumps corresponds to the hull switching to a different weight value (compare with the raw data for AwPBNF in Figure 15). All of the parallel algorithms found good solutions faster than serial AwA* and serial ARA*. Some parallel algorithms, however, took longer to prove optimality than AwA* in this domain.

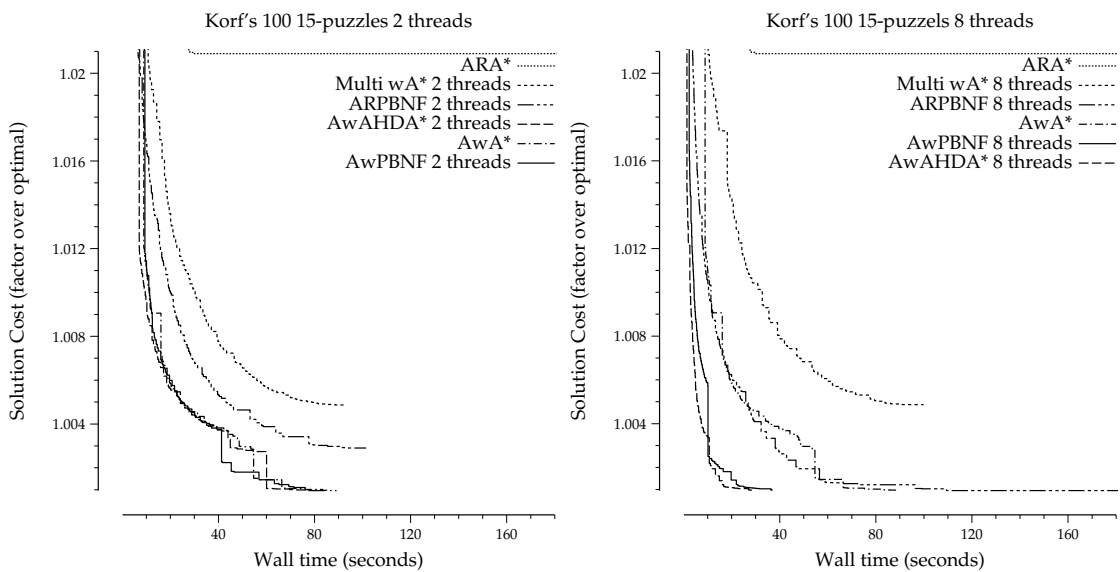

Figure 17: Korf's 100 15-puzzles lower hull anytime profiles.

## 8.2 Sliding Tile Puzzles

Figure 17 presents lower hulls for the anytime algorithms on Korf's 100 instances of the 15-puzzle. In this figure, the x-axes show the total wall clock time in seconds. These times are not normalized to A* because it is not able to solve all of the instances. In these panels, we see that AwAHDA* tended to find good solutions faster than all other algorithms. AwA* and AwPBNF performed very similarly at two threads and as the number of threads increased AwPBNF begun to find better solutions faster than AwA*. ARPBNF took longer to find good solutions than AwPBNF and AwAHDA* but it was able to find better solutions faster than its serial counterpart. The simple Multi wA* algorithm performed the worst of the parallel algorithms. Increasing the number of threads used in Multi-wA* did not seem to increase the solution quality. ARA* gave the worst performance in this domain; its profile curve can be seen at the very top of these three panels.

| | | AwAPRA* | | | | AwAHDA* | | | |
|---|---|---|---|---|---|---|---|---|---|
| | | 1.5 | 2 | 3 | 5 | 1.5 | 2 | 3 | 5 |
| **2 threads** | logistics-6 | 1.09 | 1.06 | 1.40 | 1.40 | 1.23 | 1.21 | 1.59 | 1.66 |
| | blocks-14 | 1.36 | 7.76 | 56.41 | >90.16 | 1.62 | 9.90 | 63.60 | **>110.16** |
| | gripper-7 | 0.78 | 0.77 | 0.76 | 0.75 | 1.35 | 1.33 | 1.32 | 1.33 |
| | satellite-6 | 0.77 | 0.78 | 0.78 | 0.76 | 1.26 | 1.23 | 1.24 | 1.23 |
| | elevator-12 | 0.64 | 0.67 | 0.69 | 0.70 | 1.20 | 1.19 | 1.16 | 1.17 |
| | freecell-3 | 1.37 | 1.43 | 4.61 | 1.37 | 1.66 | 1.68 | **5.65** | 1.95 |
| | depots-7 | 1.24 | 1.30 | 1.30 | 2.68 | 1.51 | 1.51 | 1.50 | 3.18 |
| | driverlog-11 | 1.15 | 1.19 | 1.11 | 1.20 | 1.50 | 1.55 | 1.46 | 1.54 |
| | gripper-8 | 0.61 | 0.62 | 0.62 | 0.62 | 1.16 | 1.11 | 1.14 | 1.11 |
| **7 threads** | logistics-6 | 1.45 | 1.43 | 1.81 | 1.81 | 2.87 | 2.81 | 3.65 | 3.74 |
| | blocks-14 | 2.54 | 15.63 | 98.52 | >177.08 | 3.30 | 19.91 | 132.97 | **>231.45** |
| | gripper-7 | 1.77 | 1.68 | 1.71 | 1.73 | 3.75 | 3.69 | 3.61 | 3.67 |
| | satellite-6 | 1.22 | 1.22 | 1.26 | 1.26 | 3.56 | 3.46 | 3.51 | 3.50 |
| | elevator-12 | 0.93 | 0.93 | 0.95 | 0.94 | 2.77 | 2.75 | 2.79 | 2.77 |
| | freecell-3 | 3.64 | 3.75 | 11.59 | 4.44 | 5.00 | 4.97 | 16.36 | **21.57** |
| | depots-7 | 3.60 | 3.64 | 3.65 | 7.60 | 4.41 | 4.42 | 4.40 | 9.25 |
| | driverlog-11 | 3.04 | 3.20 | 3.05 | 3.17 | 4.74 | 4.82 | 4.66 | 4.87 |

| | | AwPBNF | | | | AwBFPSDD | | | |
|---|---|---|---|---|---|---|---|---|---|
| | | 1.5 | 2 | 3 | 5 | 1.5 | 2 | 3 | 5 |
| **2 threads** | logistics-6 | 1.06 | 1.35 | **1.94** | **1.98** | 0.68 | 0.91 | 0.91 | 0.56 |
| | blocks-14 | 1.91 | 1.99 | 13.22 | >22.36 | 1.02 | 1.18 | 7.71 | >11.92 |
| | gripper-7 | **2.05** | **1.96** | **1.99** | **1.95** | **1.94** | **1.89** | **1.94** | 1.82 |
| | satellite-6 | 1.58 | **1.96** | **1.98** | **1.91** | **1.85** | **1.87** | 1.49 | **1.80** |
| | elevator-12 | **2.01** | 2.07 | 2.13 | 2.07 | 1.74 | 1.74 | 1.75 | 1.69 |
| | freecell-3 | 1.93 | 1.06 | 2.78 | **6.23** | 1.45 | 1.46 | 1.97 | 3.08 |
| | depots-7 | 1.94 | 2.00 | 2.01 | **4.10** | 1.44 | 1.45 | 1.32 | 2.40 |
| | driverlog-11 | **1.95** | **2.10** | **1.99** | 0.77 | 1.73 | 1.78 | 1.59 | 1.41 |
| | gripper-8 | **2.04** | **2.05** | **2.09** | **2.06** | **2.01** | **2.00** | **1.98** | **1.96** |
| **7 threads** | logistics-6 | 2.04 | 2.46 | **4.19** | **4.21** | 1.02 | 1.35 | 1.37 | 0.92 |
| | blocks-14 | 3.72 | 22.37 | 25.69 | >7.20 | 1.60 | 1.96 | 12.10 | >19.94 |
| | gripper-7 | **5.61** | **5.05** | 5.03 | **5.06** | 4.30 | 4.24 | 4.16 | 3.96 |
| | satellite-6 | **5.96** | 4.66 | 5.74 | 4.70 | 4.10 | 3.54 | 4.16 | 3.88 |
| | elevator-12 | **6.18** | **6.03** | **6.20** | **6.05** | 3.71 | 3.74 | 3.73 | 3.38 |
| | freecell-3 | 3.54 | 1.50 | 15.32 | 11.46 | 1.78 | 1.82 | 2.59 | 4.14 |
| | depots-7 | 5.74 | 5.52 | 5.48 | **10.84** | 2.02 | 1.96 | 1.92 | 3.68 |
| | driverlog-11 | **5.78** | **5.83** | **5.73** | 2.18 | 2.58 | 2.86 | 2.57 | 2.34 |

Table 6: Speed-up of anytime search to optimality over serial AwA* on STRIPS planning using various weights.

## 8.3 STRIPS Planning

Table 6 shows the speedup of the parallel anytime algorithms over serial anytime A*. All algorithms were run until an optimal solution was proved. (For a weight of 5, AwA* ran out of memory on blocks-14, so our speedup values at that weight for that instance are lower bounds.) The **bold** entries in the table represent values that are within 10% of the best performance for the given domain. For all algorithms, speedup over serial generally increased with more threads and a higher weight. PBNF gave the fastest performance for all except two domains (blocks-14 and freecell-3). In these two domains the AwAHDA* gave the best performance by at least a factor of 10x over AwPBNF.

| | | AwPBNF | | | | AwBFPSDD | | | | AwAPRA* | | |
|---|---|---|---|---|---|---|---|---|---|---|---|---|
| | | 1.5 | 2 | 3 | 5 | 1.5 | 2 | 3 | 5 | 1.5 | 2 | 3 | 5 |
| | logistics-6 | 1.48 | 1.84 | 2.36 | 2.27 | 0.68 | 0.93 | 0.71 | 0.54 | 1.12 | 1.08 | 1.08 | 0.98 |
| | blocks-14 | 1.24 | 1.22 | 0.21 | 0.03 | 0.87 | 0.18 | 0.16 | 0.16 | 1.46 | 1.46 | 1.42 | 0.94 |
| | gripper-7 | 1.07 | 0.99 | 0.99 | 1.00 | 0.93 | 0.95 | 0.93 | 0.92 | 0.99 | 1.03 | 1.01 | 0.99 |
| 7 Threads | satellite-6 | 1.10 | 0.87 | 1.08 | 0.88 | 0.88 | 0.77 | 0.91 | 0.90 | 0.99 | 1.00 | 1.01 | 1.02 |
| | elevator-12 | 1.06 | 1.04 | 1.04 | 1.03 | 0.77 | 0.78 | 0.76 | 0.73 | 1.02 | 1.00 | 1.00 | 1.00 |
| | freecell-3 | 1.05 | 0.44 | 0.99 | 0.29 | 0.64 | 0.64 | 0.20 | 0.14 | 1.13 | 1.16 | 0.82 | 0.10 |
| | depots-7 | 1.20 | 1.15 | 1.15 | 1.08 | 0.54 | 0.53 | 0.52 | 0.49 | M | M | M | M |
| | driverlog-11 | 1.16 | 1.15 | 1.19 | 0.43 | 0.53 | 0.58 | 0.54 | 0.50 | M | M | M | M |
| | gripper-8 | 1.06 | 0.99 | 0.99 | 1.00 | 0.99 | 0.98 | 0.99 | 0.97 | M | M | M | M |

Table 7: Speed-up of anytime search to optimality over PBNF on STRIPS planning problems using various weights.

Hansen and Zhou (2007) show that AwA* can lead to speedup over A* for some weight values in certain domains. Finding a suboptimal solution quickly allows $f$ pruning that keeps the open list short and quick to manipulate, resulting in faster performance even though AwA* expands more nodes than A*. We found a similar phenomenon in the corresponding parallel case. Table 7 shows speedup over unweighted optimal PBNF when using various weights for the anytime algorithms. A significant fraction of the values are greater than 1, representing a speedup when using the anytime algorithm instead of the standard optimal parallel search. In general, speedup seems more variable as the weight increases. For a weight of 1.5, AwPBNF always provides a speedup.

## 8.4 Summary

In this part of the paper we have shown how to create some new parallel anytime search algorithms based on the frameworks introduced in the previous sections. We have also created a new parallel anytime algorithm that simply runs many weighted A* searches with differing weights. In our experiments, we have seen that AwPBNF and AwAHDA* found higher quality solutions faster than other algorithms and that they both showed improved performance as more threads were added. Additionally, ARPBNF, a parallel algorithm that is based on ARA*, improved with more threads and tended to give a smoother increase in solution quality than the former two algorithms, although it did not find solutions quite as quickly and it was unable to converge on the optimal solution in the sliding tiles domain within the given time limit. Running multiple weighted A* searches did not give solutions faster as the number of threads increased, and its convergence performance was mixed.

## 9. Discussion

We have explored a set of best-first search algorithms that exploit the parallel capabilities of modern CPUs. First we looked at parallel optimal search with (Safe) PBNF, several variants of PRA* and a set of simpler previously proposed algorithms. Overall, Safe PBNF gave the best performance for optimal search. Next we created a set of bounded-suboptimal search algorithms based on PBNF, the successful variants of PRA*, and the BFPSDD algorithm. PBNF and PRA* with asynchronous communication and abstraction (AHDA*) gave the best performance over all, with PBNF doing slightly better on the average. In addition, we showed some results that suggest that bounded-suboptimal PBNF has more of an advantage over serial weighted A* search as problem difficulty increases. Finally we converted PBNF and PRA* into anytime algorithms and compared them with some serial anytime algorithms and a new algorithm called multi-weighted A*. We found that anytime weighted PBNF and the anytime variant of AHDA* gave the best anytime performance and were occasionally able to find solutions faster than their non-anytime counterparts.

Our results show that PBNF outperforms PSDD. We believe that this is because of the lack of layer-based synchronization and a better utilization of heuristic cost-to-go information. The fact that BFPSDD got better as its $f$ layers were widened is suggestive evidence. Another less obvious reason why PBNF may perform better is because a best-first search can have a larger frontier size than the breadth-first heuristic search used by PSDD. This larger frontier size will tend to create more $n$blocks containing open search nodes. There will be more disjoint duplicate detection scopes with nodes in their open lists and, therefore, more potential for increased parallelism.

Some of our results show that, even for a single thread, PBNF can outperform a serial A* search (see Table 1). This may be attributed in part to the speculative behavior of the PBNF algorithm. Since PBNF uses a minimum number of expansions before testing if it should switch to an $n$block with better $f$ values, it will search some sub-optimal nodes that A* would not search. In order to get optimal solutions, PBNF acts as an anytime algorithm; it stores incumbent solutions and prunes until it can prove that it has an optimal solution. Zhou and Hansen show that this approach has the ability to perform better than A* (Hansen & Zhou, 2007) because of upper bound pruning, which reduces the number of expansions of nodes with an $f$ value that is equal to the optimal solution cost and can reduce the number of open nodes, increasing the speed of operations on the open list. PBNF may also give good single thread performance because it breaks up the search frontier into many small open lists (one for each $n$block). Because of this, each of the priority queue operations that PBNF performs can be on much smaller queues than A*, which uses one big single queue (see Section 4.6.2).

## 9.1 Possible Extensions

While the basic guideline for creating a good abstractions in SDD (and PBNF) is to minimize the connectivity between abstract states, there are other aspects of abstraction that could be explored. For instance, discovering which features are good to include or abstract away may be helpful to users of PBNF. Too much focus on one feature could cause good nodes to be too focused in a small subset of $n$blocks (Zhou & Hansen, 2009). Likewise, size of the abstraction could be examined in more detail. Although we always use a constant abstraction size in our current work for simplicity it seems likely that abstraction size should change when number of threads changes or perhaps even based on features of the domain or problem instance. If a guideline could be devised, such as a ratio between number of $n$blocks to threads or $h$ value of the start state, a problem-adaptive abstraction size would be much simpler in real world use. Additionally, edge partitioning (Zhou & Hansen, 2007) could allow us to reduce connectivity of the abstraction used by PBNF, but further study will be necessary to discover the full impact of this technique on PBNF's behavior.

Some possible future extensions to PBNF include adaptive minimum expansion values, use of external memory, and extension to a distributed setting. Our preliminary work on adapting minimum expansion values indicated that simply increasing or decreasing based on lock failures and successes had either neutral or negative effect on performance. One reason for this may be because the minimum expansions parameter adds speculation.

It may be possible to combine PBNF with PRA* in a distributed memory setting. This algorithm may use a technique based on PRA* to distribute portions of the search space among different nodes on a cluster of work stations while using a multicore search such as PBNF on each node.

An additional technique that was not explored in this paper is running multicore search algorithms with more threads than there are available cores. This technique has been used to improve the performance of parallel delayed duplicate detection (Korf, 1993; Korf & Schultze, 2005) which is heavily I/O intensive. Using this approach, when one thread is blocked on I/O another thread can make use of the newly available processing core. Even without disk I/O this technique may be useful if threads spend a lot of time waiting to acquire locks.

## 10. Conclusions

In this paper we have investigated algorithms for best-first search on multicore machines. We have shown that a set of previously proposed algorithms for parallel best-first search can be much slower than running A* serially. We have presented a novel hashing function for PRA* that takes advantage of the locality of a search space and gives superior performance. Additionally, we have verified results presented by Kishimoto et al. (2009) that using asynchronous communication in PRA* allows it to perform better than using synchronous communication. We present a new algorithm, PBNF, that approximates a best-first search ordering while trying to keep all threads busy. We proved the correctness of the PBNF search framework and used it to derive new suboptimal and anytime algorithms.

We have performed a comprehensive empirical comparison with optimal, suboptimal and anytime variations of parallel best-first search algorithms. Our results demonstrate that using a good abstraction to distribute nodes in PRA* can be more beneficial than asynchronous communication, but that these two techniques can be used together (yielding AHDA*). We also found that the original breadth-first PSDD algorithm does not give competitive behavior without a tight upper bound for pruning. We implemented a novel extension to PSDD, BFPSDD, that gives reasonable performance on all domains we tested. Our experiments, however, demonstrate that the new PBNF and AHDA* algorithms outperformed all of the other algorithms. PBNF performs best for optimal and bounded-suboptimal search and both PBNF and AHDA* gave competitive anytime performance.

## Acknowledgments

We gratefully acknowledge support from NSF (grant IIS-0812141), the DARPA CSSG program (grant HR0011-09-1-0021) and helpful suggestions from Jordan Thayer. Some of these results were previously reported by Burns, Lemons, Zhou, and Ruml (2009b) and Burns, Lemons, Ruml, and Zhou (2009a).

## Appendix A. Pseudo-code for Safe PBNF

In the following pseudo code there are three global structures. The first is a pointer to the current incumbent solution, *incumbent*, the second is a *done* flag that is set to true when a thread recognizes that the search is complete and the third is the $n$block graph. The $n$block graph structure contains the list of free $n$blocks, *freelist* along with the $\sigma$ and $\sigma_h$ values for each $n$block. For simplicity, this code uses a single lock to access either structure. Each thread also has a local *exp* count. The *best* function on a set of $n$blocks results in the $n$block containing the open node with the lowest $f$ value.

SEARCH(INITIAL NODE)
1.  insert initial node into open
2.  for each $p \in processors$, THREADSEARCH()
3.  while threads are still running, *wait*()
4.  return *incumbent*

THREADSEARCH()
1.  $b \leftarrow$ NULL
2.  while $\neg done$
3.      $b \leftarrow$ NEXTNBLOCK($b$)
4.      $exp \leftarrow 0$
5.      while $\neg$SHOULDSWITCH($b$, $exp$)
6.          $m \leftarrow$ best open node in $b$
7.          if $m > incumbent$ then prune $m$
8.          if $m$ is a goal then
9.              if $m < incumbent$ then
10.                 lock; $incumbent \leftarrow m$; unlock
11.         else if $m$ is not a duplicate then
12.             $children \leftarrow expand(m)$
13.             for each $child \in children$
14.                 insert *child* into open of appropriate nblock
15.         $exp \leftarrow exp + 1$

SHOULDSWITCH(B, EXP)
1.  if $b$ is empty then return true
2.  if $exp < min\text{-}expansions$ then return false
3.  $exp \leftarrow 0$
4.  if $best(freelist) < b$ or $best(interferenceScope(b)) < b$ then
5.      if $best(interferenceScope(b)) < best(freelist)$ then
6.          SETHOT($best(interferenceScope(b))$)
7.      return true
8.  lock
9.  for each $b' \in interferenceScope(b)$
10.     if $hot(b')$ then SETCOLD($b'$)
11. unlock
12. return false

SETHOT(B)
1.   lock
2.   if $\neg hot(b)$ and $\sigma(b) > 0$
3.      and $\neg \exists i \in interferenceScope(b) : i < b \wedge hot(i)$ then
4.          $hot(b) \leftarrow$ true
5.          for each $m' \in interferenceScope(b)$
6.              if $hot(m')$ then SETCOLD($m'$)
7.              if $\sigma(m') = 0$ and $\sigma_h(m') = 0$
8.                  and $m'$ is not empty then
9.                      $freelist \leftarrow freelist \setminus \{m'\}$
10.             $\sigma_h(m') \leftarrow \sigma_h(m') + 1$
11.  unlock

SETCOLD(B)
1.   $hot(b) \leftarrow$ false
2.   for each $m' \in interferenceScope(b)$
3.      $\sigma_h(m') \leftarrow \sigma_h(m') - 1$
4.      if $\sigma(m') = 0$ and $\sigma_h(m') = 0$ and $m'$ is not empty then
5.          if $hot(m')$ then
6.              SETCOLD($m'$)
7.          $freelist \leftarrow freelist \cup \{m'\}$
8.          wake all sleeping threads

RELEASE(B)
1.   for each $b' \in interferenceScope(b)$
2.      $\sigma(b') \leftarrow \sigma(b') - 1$
3.      if $\sigma(b') = 0$ and $\sigma_h(b') = 0$ and $b'$ is not empty then
4.          if $hot(b')$ then
5.              SETCOLD($b'$)
6.          $freelist \leftarrow freelist \cup \{b'\}$
7.          wake all sleeping threads

NEXTNBLOCK(B)
1.   if $b$ has no open nodes or $b$ was just set to hot then lock
2.   else if $trylock()$ fails then return $b$
3.   if $b \neq$ NULL then
4.      $bestScope \leftarrow best(interferenceScope(b))$
5.      if $b < bestScope$ and $b < best(freelist)$ then
6.          unlock; return $b$
7.      RELEASE($b$)
8.   if $(\forall l \in nblocks : \sigma(l) = 0)$ and $freelist$ is empty then
9.      $done \leftarrow$ true
10.     wake all sleeping threads
11.  while $freelist$ is empty and $\neg done$, sleep
12.  if $done$ then $n \leftarrow$ NULL

13.   else
14.     $m \leftarrow best(freelist)$
15.     for each $b' \in interferenceScope(m)$
16.       $\sigma(b') \leftarrow \sigma(b') + 1$
17.  unlock
18.  return $m$

## Appendix B. TLA$^+$ Model: Hot $N$blocks

Here we present the model used to show that Safe PBNF is live-lock free. Refer to Section 3.2.3.

────────── MODULE *HotNblocks* ──────────

EXTENDS *FiniteSets*, *Naturals*

CONSTANTS *nnblocks*, *nprocs*, *search*, *nextblock*, *none*

VARIABLES *state*, *acquired*, *isHot*, *Succs*

$Vars \triangleq \langle state, acquired, isHot, Succs \rangle$

$States \triangleq \{ search, nextblock \}$

$Nblocks \triangleq 0 \,..\, nnblocks - 1$

$Procs \triangleq 0 \,..\, nprocs - 1$

ASSUME $nnblocks \geq nprocs \wedge nprocs > 0 \wedge nnblocks > 1 \wedge none \notin Nblocks \wedge Cardinality(States) = 2$

$Preds(x) \triangleq \{ y \in Nblocks : x \in Succs[y] \}$    Set of predecessors to *Nblock x*

$IntScope(x) \triangleq Preds(x) \cup \text{UNION} \{ Preds(y) : y \in Succs[x] \}$    The interference scope of *x*

$IntBy(x) \triangleq \{ y \in Nblocks : x \in IntScope(y) \}$    Set of *Nblocks* which *x* interferes.

$Busy(A) \triangleq A \cup \text{UNION} \{ Succs[x] : x \in A \}$    Set of *Nblocks* which are busy given the set of acquired nblocks

$Overlap(x, A) \triangleq A \cap IntScope(x)$    Set of *Busy Nblocks* overlapping the successors of *x*

$Hot(A) \triangleq \{ x \in Nblocks : isHot[x] \wedge Overlap(x, A) \neq \{\} \}$    Set of all hot nblocks given the set of acquired nblocks

$HotInterference(A) \triangleq \text{UNION} \{ IntScope(x) : x \in Hot(A) \}$    Set of *Nblocks* in interference scopes of hot nblocks

$Free(A) \triangleq \{ x \in Nblocks : Overlap(x, A) = \{\} \wedge x \notin HotInterference(A) \}$    Free *Nblocks*

$Acquired \triangleq \{ acquired[x] : x \in Procs \} \setminus \{ none \}$    Set of *Nblocks* which are currently acquired

$OverlapAmt(x) \triangleq Cardinality(Overlap(x, Acquired))$    The number of nblocks overlapping *x*.

$doNextBlock(x) \triangleq \wedge$ UNCHANGED $\langle Succs \rangle$

$\wedge state[x] = nextblock \wedge acquired[x] = none \Rightarrow Free(Acquired) \neq \{\}$

$\wedge$ IF $Free(Acquired \setminus \{ acquired[x] \}) \neq \{\}$ THEN

     $\wedge \exists y \in Free(Acquired \setminus \{ acquired[x] \}) : acquired' = [acquired \text{ EXCEPT } ![x] = y]$

     $\wedge state' = [state \text{ EXCEPT } ![x] = search]$

     $\wedge isHot' = [y \in Nblocks \mapsto \text{IF } y \in Free(Acquired \setminus \{ acquired[x] \})$

                       THEN FALSE ELSE $isHot[y]]$

ELSE   $\wedge acquired' = [acquired \text{ EXCEPT } ![x] = none]$

        $\wedge isHot' = [y \in Nblocks \mapsto \text{IF } y \in Free(Acquired')$

                       THEN FALSE ELSE $isHot[y]]$

        $\wedge$ UNCHANGED $\langle state \rangle$

$doSearch(x) \triangleq \wedge$ UNCHANGED $\langle acquired, Succs \rangle$

        $\wedge state[x] = search \wedge state' = [state \text{ EXCEPT } ![x] = nextblock]$

        $\wedge \vee$ UNCHANGED $\langle isHot \rangle$

            $\vee \exists y \in IntBy(acquired[x]) : \wedge \neg isHot[y]$

                                 $\wedge IntScope(y) \cap Hot(Acquired) = \{\}$

                                 $\wedge y \notin HotInterference(Acquired)$

                                 $\wedge isHot' = [isHot \text{ EXCEPT } ![y] = \text{TRUE}]$

$Init \triangleq \wedge state = [x \in Procs \mapsto nextblock] \wedge acquired = [x \in Procs \mapsto none]$

      $\wedge isHot = [x \in Nblocks \mapsto \text{FALSE}]$

      This is a basic graph where each nblock is connected to its neighbors forming a loop.

      $\wedge Succs = [x \in Nblocks \mapsto \text{IF } x = 0 \text{ THEN } \{ nnblocks - 1, x + 1 \}$

                            ELSE IF $x = nnblocks - 1$ THEN $\{ 0, x - 1 \}$ ELSE $\{ x - 1, x + 1 \}]$

$Next \triangleq \exists x \in Procs : (doNextBlock(x) \vee doSearch(x))$

$Fairness \triangleq \forall x \in Procs : \text{WF}_{Vars}(doNextBlock(x) \vee doSearch(x))$

$Prog \triangleq Init \wedge \Box[Next]_{Vars} \wedge Fairness$

$HotNblocks \triangleq \forall x \in Nblocks : isHot[x] \rightsquigarrow \neg isHot[x]$    The property to prove